\documentclass{article}

\usepackage{iclr2021_conference,times}
\usepackage{amsmath, amsfonts}
\usepackage{multirow}
\usepackage{mathtools}
\usepackage{xspace}
\usepackage{enumerate}
\usepackage[ruled,vlined]{algorithm2e}
\usepackage{wrapfig}
\usepackage{paralist}
\usepackage[pagebackref=true,breaklinks=true,letterpaper=true,colorlinks,bookmarks=false,citecolor=blue]{hyperref}
\usepackage{graphicx}
\usepackage{booktabs}% for \midrule and \cmidrule macros
\iclrfinalcopy
\usepackage{amsmath,amsfonts,bm}

\def\eqref#1{equation~\ref{#1}}
\def\ceil#1{\lceil #1 \rceil}

\def\1{\bm{1}}

\DeclareMathAlphabet{\mathsfit}{\encodingdefault}{\sfdefault}{m}{sl}
\SetMathAlphabet{\mathsfit}{bold}{\encodingdefault}{\sfdefault}{bx}{n}

\DeclareMathOperator*{\argmax}{arg\,max}

\title{Hierarchical Reinforcement Learning by \\ Discovering Intrinsic Options}

\newcommand{\method}{HIDIO\xspace} % placeholder; to make a name for the method
\newcommand{\pihi}{\ensuremath{\pi_{\theta}}}
\newcommand{\pilow}{\ensuremath{\pi_{\phi}}}
\newcommand{\eg}{\textit{e.g.},}

\newcommand{\bfs}{\mathbf{s}}
\newcommand{\bfa}{\mathbf{a}}
\newcommand{\bfu}{\mathbf{u}}
\newcommand{\mbfs}{\overline{\mathbf{s}}}
\newcommand{\mbfa}{\overline{\mathbf{a}}}
\newcommand{\expect}{\mathop{\mathbb{E}}}
\newcommand{\defeq}{\vcentcolon=}
\newcommand{\framework}{HIDIO}
\newcommand{\smtexttt}[1]{{\small\texttt{#1}}}

\newcommand{\haonan}[1]{{\color{red}Haonan: #1}}
\newcommand{\jesse}[1]{{\color{blue}Jesse: #1}}

\author{Jesse Zhang\thanks{Denotes equal contribution. Email to \texttt{jessez@usc.edu, \{haonan.yu,wei.xu\}@horizon.ai}} \ \thanks{Work done as an intern at Horizon Robotics.} \ \textsuperscript{1}, Haonan Yu\footnotemark[1] \ \textsuperscript{2}, Wei Xu\textsuperscript{2}\\
\textsuperscript{1}University of Southern California, \textsuperscript{2}Horizon Robotics
\\

}

\begin{document}
\maketitle

%===============================================================================
\begin{abstract}
%(\textbf{HI}erarchical rl by \textbf{D}iscovering \textbf{I}ntrinsic \textbf{O}ptions)
We propose a hierarchical reinforcement learning method, \method, that can learn
task-agnostic options in a self-supervised manner while jointly learning to utilize 
them to solve sparse-reward tasks. Unlike current hierarchical RL approaches
that tend to formulate goal-reaching low-level tasks or pre-define ad hoc lower-level policies,
\method\ encourages lower-level option learning that is independent of the
task at hand, requiring few assumptions or little knowledge about the task structure.
These options are learned through an intrinsic entropy minimization objective conditioned
on the option sub-trajectories. The learned options are diverse and task-agnostic. In experiments on sparse-reward robotic manipulation and navigation tasks, \method\ achieves higher success rates with greater sample efficiency than 
regular RL baselines and two state-of-the-art hierarchical RL methods. Code available at \url{https://www.github.com/jesbu1/hidio}.
\end{abstract}

%!TEX root=main.tex
\section{Introduction}
\label{sec:introduction}
%===============================================================================
Imagine a wheeled robot learning to kick a soccer ball into a goal with sparse
reward supervision. In order to succeed, it must discover how to first navigate in its environment,
then touch the ball, and finally kick it into the goal, only receiving a positive reward at the end for completing the task. This is a naturally difficult problem for traditional reinforcement learning (RL)
to solve, unless the task has been manually decomposed into temporally extended stages where each 
stage constitutes a much easier subtask. In this paper we ask, how do we learn to decompose 
the task automatically and utilize the decomposition to solve sparse reward problems?

Deep RL has made great strides solving a variety of tasks recently, with
hierarchical RL (hRL) demonstrating promise in solving such sparse reward
tasks 
\citep{sharma2018directedinfo, le2018hierarchical, merel2018hierarchical,
ranchod2015nonparametric}. 
In hRL, the task is decomposed into a hierarchy of subtasks, where policies at the top of the hierarchy call upon policies below to perform actions to solve their respective subtasks. This abstracts away actions for the policies at the top levels of the hierarchy. 
hRL makes exploration easier
by potentially reducing the number of steps the agent needs to take
to explore its state space. Moreover, at higher levels of the hierarchy, temporal abstraction results 
in more aggressive, multi-step value bootstrapping when temporal-difference (TD) 
learning is employed. These benefits are critical in sparse
reward tasks as they allow an agent to more easily discover reward signals and assign credit.

Many existing hRL methods make assumptions about the task structure (\eg\ fetching an object involves three stages: moving
towards the object, picking it up, and combing back), and/or the skills needed to solve the task (\eg\ pre-programmed motor skills)
\citep{florensa2017stochastic, Riedmiller2018, lee2018composing,
hausman2018learning,Lee2020Learning,sohn2018hierarchical,ghavamzadeh2003hierarchical, nachum2018dataefficient}. Thus these methods may require manually designing the correct
task decomposition, explicitly formulating the option space, or programming pre-defined options for higher level policies to compose. 
Instead, we seek to formulate a general method that can learn these abstractions from scratch, for any task, 
with little manual design in the task domain.

%The options framework \citep{Sutton1999} 
%However, the classical option
%definition consists of an option-dependent policy, initial
%state set, and option termination conditions, all of which are usually specific to 
%each environment and task. 

The main contribution of this paper is \framework\ (\textbf{HI}erarchical RL
by \textbf{D}iscovering \textbf{I}ntrinsic \textbf{O}ptions), a hierarchical method that discovers
task-agnostic intrinsic options in a self-supervised manner while learning to schedule them to accomplish environment tasks. 
The latent option representation is uncovered as the option-conditioned policy is trained, 
both according to the same self-supervised worker objective. The scheduling of options is simultaneously learned by maximizing environment
reward collected by the option-conditioned policy.
HIDIO can be easily applied to new sparse-reward tasks by simply re-discovering options.
We propose and empirically evaluate various instantiations of the option discovery process, comparing the resulting 
options with respect to their final task performance. We demonstrate that \framework\ is 
able to efficiently learn and discover diverse options to be utilized 
for higher task reward with superior sample efficiency compared to other hierarchical 
 methods.
%===============================================================================
%!TEX root=main.tex
\section{Preliminaries}
\label{sec:preliminaries}
We consider the reinforcement learning (RL) problem in a Markov Decision
Process (MDP). Let $\mathbf{s}\in \mathbb{R}^S$ be the agent state. We use
the terms ``state'' and ``observation'' interchangeably to denote the
environment input to the agent. A state can be fully or partially observed.
Without loss of generality, we assume a continuous action space
$\mathbf{a}\in \mathbb{R}^A$ for the agent. Let
$\pi_{\theta}(\mathbf{a}|\mathbf{s})$ be the policy distribution with
learnable parameters $\theta$, and
$\mathcal{P}(\mathbf{s}_{t+1}|\mathbf{s}_t,\mathbf{a}_t)$ the transition
probability that measures how likely the environment transitions to
$\mathbf{s}_{t+1}$ given that the agent samples an action by $\mathbf{a}_t
\sim \pi_{\theta}(\cdot|\mathbf{s}_t)$. After the transition to
$\mathbf{s}_{t+1}$, the agent receives a deterministic scalar reward $r(\mathbf{s}_t, \mathbf{a}_t, \mathbf{s}_{t+1})$.
%\footnote{We assume a
%deterministic reward function for simplicity. Our discussion in
%this paper also applies to a stochastic reward function.} 

The objective of RL is to maximize the sum of discounted rewards with respect to
$\theta$: 
\begin{equation}
\label{eq:rl_objective}
    \expect_{\pi_{\theta},\mathcal{P}}
    \left[ \sum_{t=0}^{\infty} \gamma^t r(\mathbf{s}_t, \mathbf{a}_t,
    \mathbf{s}_{t+1})\right]
\end{equation}
where $\gamma \in [0, 1]$ is a discount factor. %Because the dynamics are 
%fixed for a given environment, we will omit $\mathcal{P}$ in the above
%expectation operator for notatio
We will omit $\mathcal{P}$ in the 
expectation for notational simplicity.

\iffalse
\textbf{Goal-conditioned RL} \ A multi-goal RL setting (\eg\
\citep{eysenbach2018diversity, plappert2018multigoal}) usually assumes that,
at the beginning of each episode $t=0$, a goal $\mathbf{u} \in \mathbb{R}^D$
is sampled from some distribution $p(\mathbf{u})$ and keeps
unchanged throughout the entire episode. As a result, both the policy
distribution $\pi_{\theta}(\mathbf{a}|\mathbf{s},\mathbf{u})$ and reward function $r_{\mathbf{u}}$, depend
on the sampled goal, and the expectation in Eq.~\ref{eq:rl_objective} is
taken over $p(\mathbf{u})$.
\fi

In the options framework~\citep{Sutton1999}, the agent can
switch between different options during an episode, where an option is
translated to a sequence of actions by an option-conditioned policy with a
termination condition. A set of options defined over an MDP induces a
hierarchy that models temporal abstraction. 
For a typical two-level hierarchy, a higher-level policy produces options, and the policy 
at the lower level outputs environment actions conditioned on the proposed 
options. The expectation in Eq.~\ref{eq:rl_objective} is taken over policies at both levels. 

%Prior works have used the terms
%``goal,'' ``option,'' and ``skill'' in similar ways, but in this paper we will use ``option.''

%===============================================================================
%!TEX root=main.tex
\section{Hierarchical RL by Discovering Intrinsic Options}
\label{sec:method}
%
\iffalse
\haonan{Should move this paragraph to an earlier section.}
%
One notorious difficulty in RL tasks with extremely sparse rewards and a long
time horizon is that the reward signal is too delayed to drive model
learning. %
This poses the challenging credit assignment problem as it takes time to
propagate meaningful learning signal back to the action that actually
contributes to meaningful events.
\fi

%\begin{wrapfigure}{R}{0.5\textwidth}
%% edit at: https://docs.google.com/presentation/d/1XfTyQT4mFDw8hpQE8bBFmMkMi5LBiOCmeTP2uMHxhFo/edit?usp=sharing
%    \centering
%    \includegraphics[width=0.5\textwidth]{images/framework_colored.pdf}
%    \caption{The overall framework of \method. Solid lines denote rollout
%    sampling while dashed lines denote training. Light-color lines denote a
%    frequency of $\frac{1}{K}$, where $K$ is the interval between two options. Refer to
%    Eq.~\ref{eq:sampling} for sampling and Eqs.~\ref{eq:scheduler_objective} and~\ref{eq:worker_objective} for
%    training.}
%    \label{fig:framework}
%\end{wrapfigure}
\begin{wrapfigure}{R}{0.55\textwidth}
    \centering
    \includegraphics[width=0.55\textwidth]{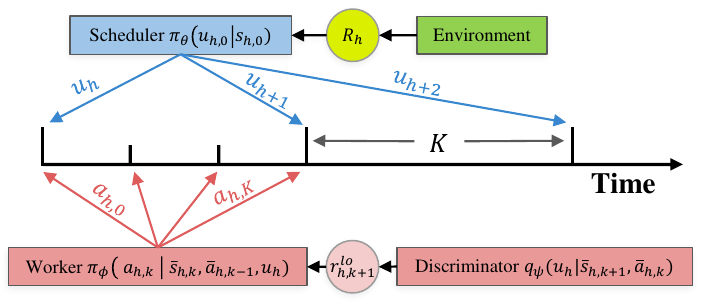}
    \caption{The overall framework of \method. The scheduler $\pi_\theta$ samples 
    an option $\bfu_h$ every $K$ (3 in this case) time steps, which is used to guide
    the worker $\pi_\phi$ to directly interact in the environment conditioned on $\bfu_h$ and the current sub-trajectory $\mbfs_{h, k}, \mbfa_{h, k-1}$. The scheduler receives accumulated environment rewards $R_h$, while the worker receives intrinsic rewards $r_{h, k+1}^{\text{lo}}$. Refer to
    Eq.~\ref{eq:sampling} for sampling and Eqs.~\ref{eq:scheduler_objective} and~\ref{eq:worker_objective} for
    training.} 
    \label{fig:framework}
    \vspace{-2pt}
\end{wrapfigure}

We now introduce our hierarchical method for solving sparse reward tasks. We assume 
little prior knowledge about the task structure, except that it can be learned through a hierarchy of two levels.
\iffalse
We allow the reward function to be defined in any form, not
necessarily defined by an ``$\epsilon$-region'' in the state space as assumed
in some sparse reward
settings~\citep{andrychowicz2017hindsight,Riedmiller2018}. 
\fi
The higher-level policy
(the \textit{scheduler} \pihi), is trained to maximize environment reward,
while the lower-level policy (the \textit{worker} \pilow) is trained in a
self-supervised manner to efficiently discover options that are utilized by
\pihi\ to accomplish tasks. Importantly, by self-supervision the worker gets access to dense
intrinsic rewards regardless of the sparsity of the extrinsic rewards.

%\subsection{Overall Hierarchy}
%
Without loss of generality, we assume that each episode has a length of $T$
and the scheduler outputs an option every $K$ steps. The scheduled option $\bfu \in [-1, 1]^D$ (where $D$ is a pre-defined dimensionality), is a latent 
representation that will be learned from scratch given the environment task. Modulated by $\bfu$, the worker executes $K$
steps before the scheduler outputs the next option.  Let the time horizon of
the scheduler be $H=\ceil{\frac{T}{K}}$. Formally, we
define

\begin{equation}
\label{eq:sampling}
    \begin{array}{lll}
        \text{Scheduler policy:} & \bfu_h \sim \pihi(\cdot|\bfs_{h,0}),&
        0\leq h < H\\
        \text{Worker policy:} & \bfa_{h,k} \sim
        \pilow(\cdot|\bfs_{h,0}, \bfa_{h, 0}, ...,\bfs_{h, k},\bfu_h),& 0 \leq k < K\\
        \text{Environment dynamics:} & \bfs_{h, k+1} \sim
        \mathcal{P}(\cdot|\bfs_{h, k},\bfa_{h, k}),& 0 \leq h < H, 0 \le k < K\\
    \end{array}
\end{equation}
\noindent where we denote $\bfs_{h,k}$ and $\bfa_{h,k}$ as the $k$-th state and action 
respectively, within the $h$-th option window of length $K$.
Note that given this sampling process, we have
$\bfs_{h,K}\equiv\bfs_{h+1,0}$, namely, the last state of the current option
$\bfu_h$ is the initial state of the next option $\bfu_{h+1}$. 
%Finally, 
%$\mbfs_{h, k}$ and $\mbfa_{h, k-1}$ denotes state and action subtrajectories respectively---
%we further explain 
%subtrajectory notation in Section~\ref{sec:learning the worker}. 
The overall
framework of our method is illustrated in Figure~\ref{fig:framework}. 

\subsection{Learning the Scheduler}
Every time the scheduler issues an option $\bfu_h$, it receives an reward
$R_h$ computed by accumulating environment rewards over the next $K$ steps.
Its objective is:

\begin{equation}
\label{eq:scheduler_objective}
%\begin{array}{ll}
%    &\displaystyle\max_{\theta}\ \mathbb{E}_{\pihi}\left[\sum_{h=0}^{H-1}\beta^h R_h\right], \\
%    \text{where}& \beta=\gamma^K\ \text{and}\ R_h=\mathbb{E}_{\pilow}\left[\sum_{k=0}^{K-1}\gamma^k r(\bfs_{h,k}, \bfa_{h,k}, \bfs_{h,k+1})\right]\\
%\end{array}
    \displaystyle\max_{\theta}\ \mathbb{E}_{\pihi}\left[\sum_{h=0}^{H-1}\beta^h R_h\right],
    \text{where}\ \ \beta=\gamma^K\ \text{and}\ R_h=\mathbb{E}_{\pilow}\left[\sum_{k=0}^{K-1}\gamma^k r(\bfs_{h,k}, \bfa_{h,k}, \bfs_{h,k+1})\right]
\end{equation}

This scheduler objective itself is not a new concept, as similar ones have
been adopted by other hRL methods
\citep{Vezhnevets2017,nachum2018dataefficient,Riedmiller2018}. One
significant difference between our option with that of prior work is that our option $\bfu$ is 
simply a latent variable; there is no explicit constraint on what semantics $\bfu$ could represent. 
In contrast, existing
methods usually require their options to reside in a subspace of the state
space, to be grounded to the environment, or to have known structures, so
that the scheduler can compute rewards and termination conditions for the
worker. Note that our latent options can be easily re-trained given a new task.

%As a result, their options are task-dependent and might have
%difficulty adapting to new scenarios without being redesigned.

%While \citet{Li2020} also formulates task-agnostic options,
%they do it in a discrete space.

\subsection{Learning the Worker}
\label{sec:learning the worker}
The main focus of this paper is to investigate how to effectively learn the
worker policy in a self-supervised manner.
%
\iffalse
Although the latter formulation is in principle very general, its successful
application to a new scenario usually asks for manually redesigning the goal
space given the new state space, or learning the goal space first if a latent
state space has to be learned from high-dimensional observations~\citep{?}.
\fi
%
Our motivation is that it might be unnecessary to make an option dictate the worker
to reach some ``$\epsilon$-space'' of goals \citep{Vezhnevets2017,nachum2018dataefficient}. As long as the option can be
translated to a short sequence of primitive actions, it does not need to be
grounded with concrete meanings such as goal reaching. 
Below we will treat the option as a latent variable that modulates the
worker, and propose to learn its latent representation in a hierarchical setting 
from the environment task.

\iffalse
Our worker learning is inspired by similar
ideas explored by DIAYN \citep{eysenbach2018diversity} and DADS
\citep{Sharma2019} for unsupervised skill discovery, but their goals are
fixed over entire episodes. Also different from them, our worker goes beyond
using individual state features alone for identifying options.
\fi

\subsubsection{Worker Objective}
We first define a new meta MDP on top of the original task MDP so that for any $h$, $k$, and $t$:

\begin{compactenum}[1)]
    \item $\mbfs_{h,k}\defeq(\bfs_{h,0},\ldots,\bfs_{h,k})$,
    \item $\mbfa_{h,k}\defeq(\bfa_{h,0},\ldots,\bfa_{h,k})$,
    \item $r(\mbfs_{h, k},\mbfa_{h, k},\mbfs_{h, k+1})\defeq r(\bfs_{h, k},\bfa_{h, k},\bfs_{h, k+1})$, and 
    \item $\mathcal{P}(\mbfs_{h, k+1}|\mbfs_{h, k},\mbfa_{h, k})\defeq \mathcal{P}(\bfs_{h, k+1}|\bfs_{h, k},\bfa_{h, k})$.
\end{compactenum}

This new MDP equips the worker with historical state and action information
since the time $(h,0)$ when an option $h$ was scheduled. Specifically, each state 
$\mbfs_{h,k}$ or action $\mbfa_{h,k}$ encodes the history from the beginning $(h,0)$ up to $(h,k)$ 
within the option. In the following, we will call pairs
$\{\mbfa_{h,k}, \mbfs_{h,k+1}\}$ option \emph{sub-trajectories}. The worker
policy now takes option sub-trajectories as inputs: $\bfa_{h,k} \sim
\pilow(\cdot|\mbfs_{h,k},\mbfa_{h,k-1},\bfu_h), 0 \leq k < K$, whereas the
scheduler policy still operates in the original MDP.

Denote $\sum_{h,k}\equiv \sum_{h=0}^{H-1}\sum_{k=0}^{K-1}$ for simplicity.
The worker objective, defined on this new MDP, is to minimize the entropy of
the option $\bfu_h$ conditioned on the option sub-trajectory
$\{\mbfa_{h,k},\mbfs_{h,k+1}\}$:

\begin{equation}
\label{eq:formulation}
    \displaystyle\max_{\phi}\ \expect_{\pihi,\pilow}\sum_{h,k}\underbrace{\log p(\bfu_h|\mbfa_{h,k},\mbfs_{h,k+1})}_{\text{negative conditional option entropy}}\underbrace{-\beta\log\pilow(\bfa_{h,k}|\mbfs_{h,k},\mbfa_{h,k-1},\bfu_h)}_{\text{worker policy entropy}}
\end{equation}
where the expectation is over the current $\pihi$ and $\pilow$ but the
maximization is only with respect to $\phi$. Intuitively, the first term
suggests that the worker is optimized to confidently identify an option given
a sub-trajectory. However, it alone will not guarantee the diversity of
options because potentially even very similar sub-trajectories can be classified
into different options if the classification model has a high capacity, in which case
we say that the resulting sub-trajectory space has a very high
``resolution''. As a result, the conditional entropy alone might not be able to
generate useful options to be exploited by the scheduler for task solving, because the coverage 
of the sub-trajectory space is poor. To
combat this degenerate solution, we add a second term which maximizes the
entropy of the worker policy. Intuitively, while the worker generates
identifiable sub-trajectories corresponding to a given option, it should act
as randomly as possible to separate sub-trajectories of different options,
lowering the ``resolution'' of the sub-trajectory space to encourage
its coverage. 

Because directly estimating the posterior
$p(\bfu_h|\mbfa_{h,k},\mbfs_{h,k+1})$ is intractable, we approximate it with
a parameterized posterior $\log q_{\psi}(\bfu_h|\mbfa_{h,k},\mbfs_{h,k+1})$
to obtain a lower bound \citep{Barber2003}, where $q_{\psi}$ is a
\emph{discriminator} to be learned. Then we can maximize this lower bound
instead:

\begin{equation}
\label{eq:worker_objective}
\displaystyle\max_{\phi,\psi}\expect_{\pihi,\pilow}\ \sum_{h,k} \log
q_{\psi}(\bfu_h|\mbfa_{h,k},\mbfs_{h,k+1})
-\beta\log\pilow(\bfa_{h,k}|\mbfs_{h,k},\mbfa_{h,k-1},\bfu_h).
\end{equation}

The discriminator $q_{\psi}$ is trained by maximizing likelihoods of options given
sampled sub-trajectories. The worker $\pilow$ is trained via max-entropy RL (Soft
Actor-Critic (SAC)~\citep{haarnoja2018soft}) with the intrinsic reward
$r^{lo}_{h,k+1} \defeq \log q_{\psi}(\cdot)-\beta\log\pilow(\cdot)$. $\beta$
is fixed to $0.01$ in our experiments. %a constant weight (fixed to $0.01$ in our experiments).

Note that there are at least four differences between
Eq.~\ref{eq:worker_objective} and the common option discovery objective in
either VIC~\citep{Gregor2016} or DIAYN~\citep{eysenbach2018diversity}:

\begin{compactenum}
    \item Both VIC and DIAYN assume that a sampled option will last through an entire
    episode, and the option is always sampled at the beginning of an episode.
    Thus their option trajectories ``radiate'' from the initial state set. In
    contrast, our worker policy learns options that initialize every $K$ steps
    within an episode, and they can have more diverse semantics depending on
    the various states $s_{h,0}$ visited by the agent. This is especially helpful for some
    tasks where new options need to be discovered after the agent reaches unseen 
    areas in later stages of training.
    \item Actions taken by the worker policy under the current option will 
    have consequences on the next option. This is because the final state 
    $s_{h,K}$ of the current option is defined to be the initial state $s_{h+1,0}$ 
    of the next option. So in general, the worker policy is trained not only 
    to discover diverse options across the current $K$ steps, but also to make 
    the discovery easier in the future steps. In other words, the worker policy needs to solve
    the credit assignment problem \emph{across} options, under the expectation of the scheduler policy.
    \item To enable the worker policy to learn from a discriminator that predicts based
    on option sub-trajectories $\{\mbfa_{h,k},\mbfs_{h,k+1}\}$ instead of solely
    on individual states $\bfs_{h,k}$, we have constructed a new meta MDP where 
    each state $\mbfs_{h,k}$ encodes history from the beginning $(h,0)$ 
    up to $(h,k)$ within an option $h$. This new meta MDP is critical, because otherwise
    one simply cannot learn a worker policy from a reward function that is defined 
    by multiple time steps (sub-trajectories) since the learning problem is no longer Markovian. 
    \item Lastly, thanks to the new MDP, we are able to explore various possible instantiations 
    of the discriminator (see Section~\ref{sec:discriminator}). As 
    observed in the experiments, individual states are actually not the optimal 
    features for identifying options.
\end{compactenum}

These differences constitute the major novelty of our worker objective.

\subsubsection{Shortsighted Worker}
It's challenging for the worker to
accurately predict values over a long horizon, since its rewards are densely
computed by a complex nonlinear function $q_{\psi}$. Also each option only
lasts at most $K$ steps. Thus we set the discount $\eta$ for the worker in
two shortsighted ways:

\begin{compactenum}
    \item \texttt{Hard}: setting $\eta=0$ every $K$-th step and $\eta=1$ otherwise.
    Basically this truncates the temporal correlation (gradients) between
    adjacent options. Its benefit might be 
    faster and easier value learning because the value is
    bootstrapped over at most $K$ steps ($K\ll T$). 
    %The downside, however, is that the
    %worker might become too short-sighted so that it leads to some state
    %$\bfs_{h+1,0}$ that makes learning future options
    %difficult (\eg\, pushing an object off a table).
    \item \texttt{Soft}: $\eta=1-\frac{1}{K}$, which considers rewards of
    roughly $K$ steps ahead. The worker policy still needs to
    take into account the identification of future option sub-trajectories,
    but their importance quickly decays.
\end{compactenum}

We will evaluate both versions and compare their performance in 
Section~\ref{sec:worker_ablations}.

\subsection{Instantiating the Discriminator} 
\label{sec:discriminator}
We explore various ways of instantiating the
discriminator $q_{\psi}$ in order to compute useful intrinsic rewards for
the worker. Previous work has utilized individual states
\citep{eysenbach2018diversity, jabri2019unsupervised} or full observation
trajectories \citep{warde-farley2018, Sharma2019,achiam2018variational} for
option discrimination. Thanks to the newly defined meta MDP, our discriminator is able to take option sub-trajectories
instead of current individual states for prediction. In this paper, we investigate
six sub-trajectory feature extractors $f_{\psi}$:
\begin{table}[!h]
    \centering
    \begin{tabular}{l|l|l|l}
         Feature extractor & Name & Formulation & Explanation \\
         \hline
         \multirow{7}{*}{$f_{\psi}(\mbfa_{h,k},\mbfs_{h,k+1})=$} & \smtexttt{State} & $\textsc{Mlp}(\bfs_{h,k+1})$ &  Next state alone \\
         %& & & (DIAYN \citep{eysenbach2018diversity})\\
         & \smtexttt{Action} & $\textsc{Mlp}([\bfs_{h,0},\bfa_{h,k}])$ & Action in context\\
         & \smtexttt{StateDiff} & $\textsc{Mlp}(\bfs_{h,k+1} - \bfs_{h,k})$ & Difference between state pairs\\
         & \smtexttt{StateAction} & $\textsc{Mlp}([\bfa_{h,k}, \bfs_{h,k+1}])$ & Action and next state\\
         & \smtexttt{StateConcat} & $\textsc{Mlp}([\mbfs_{h,k+1}])$ & Concatenation of states\\
         & \smtexttt{ActionConcat} & $\textsc{Mlp}([\bfs_{h,0}, \mbfa_{h,k}])$ & Concatenation of actions\\
    \end{tabular}
    \label{tab:features}
\end{table}

where the operator $[\cdot]$ denotes concatenation and \textsc{Mlp} denotes a
multilayer perceptron\footnote{In this paper we focus on non-image
observations that can be processed with MLPs, although our method doesn't
have any assumption about the observation space.}.
%When using actions
%for prediction, we always condition them on the first state of the option, so
%potentially the same option can map to a different policy depending on the
%current context. \jesse{$\leftarrow$ Not sure what this sentence exactly means}
Our \texttt{State} feature extractor is most similar to DIAYN
\citep{eysenbach2018diversity}, and \texttt{StateConcat} is similar to
\citep{warde-farley2018, Sharma2019,achiam2018variational}. However we note
that unlike these works, the distribution of our option sub-trajectories is
also determined by the scheduler in the context of hRL. The other four feature
extractors have not been evaluated before.
With the extracted feature, the log-probability of predicting an option is
simply computed as the negative squared L2 norm:
$\log q_{\psi}(\bfu_h|\mbfa_{h,k},\mbfs_{h,k+1}) = -\Vert
f_{\psi}(\mbfa_{h,k},\mbfs_{h,k+1}) - \bfu_h\Vert_2^2$,
by which we implicitly assume the discriminator's output distribution to be
a $\mathcal{N}(\bf0, I_D)$ multivariate Gaussian. %with zero means and unit variances.

%The benefit of
%this is that rewards for the worker are always upper bounded by 0.

\subsection{Off-policy Training}

The scheduler and worker objectives (Eq.~\ref{eq:scheduler_objective} and
Eq.~\ref{eq:worker_objective}) are trained jointly. In principle, on-policy
training such as A2C \citep{Clemente17} is needed due to the interplay
between the scheduler and worker. However, to reuse training data and improve
sample efficiency, we employ off-policy training
(SAC~\citep{haarnoja2018soft}) for both objectives with some modifications.

\textbf{Modified worker objective}\ In practice, the expectation over the
scheduler $\pihi$ in Eq.~\ref{eq:worker_objective} is replaced with the
expectation over its historical versions. Specifically, we sample options
$\bfu_h$ from a replay buffer, together with sub-trajectories
$\{\mbfa_{h,k},\mbfs_{h,k+1}\}$. This type of data distribution modification
is conventional in off-policy training \citep{lillicrap2015continuous}.

\textbf{Intrinsic reward relabeling}\ 
%Because the intrinsic rewards in the replay 
%buffer for training the worker policy become outdated once the discriminator updates, 
%
We always recompute the rewards in Eq.~\ref{eq:worker_objective} using the
up-to-date discriminator for every update of $\phi$, which can be
trivially done without any additional interaction with the environment.

\textbf{Importance correction}\ The data in the replay buffer was generated
by historical worker policies. Thus a sampled option sub-trajectory will be
outdated under the same option, causing confusion to the scheduler policy. To
resolve this issue, when minimizing the temporal-difference (TD) error
between the values of $\bfs_{h,0}$ and $\bfs_{h+1,0}$ for the scheduler, an
importance ratio can be multiplied:
$\prod_{k=0}^{K-1}\frac{\pilow(\bfa_{h,k}|\mbfs_{h,k},\mbfa_{h,k-1},\bfu_h)}{\pilow^{old}(\bfa_{h,k}|\mbfs_{h,k},\mbfa_{h,k-1},\bfu_h)}$.
A similar correction can also be applied to the discriminator
loss. However, in practice we find that this ratio has a very high variance
and hinders the training. Like the similar observations
made in~\citet{nachum2018dataefficient,fedus2020revisiting}, even without importance correction our method is able to perform well empirically\footnote{One possible reason is that the deep RL process is
``highly non-stationary anyway, due to changing policies, state distributions
and bootstrap targets''~\citep{schaul2015prioritized}.}.

%===============================================================================
%!TEX root=main.tex 
\section{Experiments} 
\label{sec:result}

\iffalse
In this section, we evaluate \method's performance on sparse reward robotic
manipulation and navigation tasks, comparing to several baselines and two
state-of-the-art hierarchical methods. We also ablate various aspects of
\method, identifying how they contribute to overall performance and task
sample efficiency.
\fi

\begin{figure}[!t]
    \centering
    \includegraphics[trim=275 0 125 0,clip,height=0.115\textwidth]{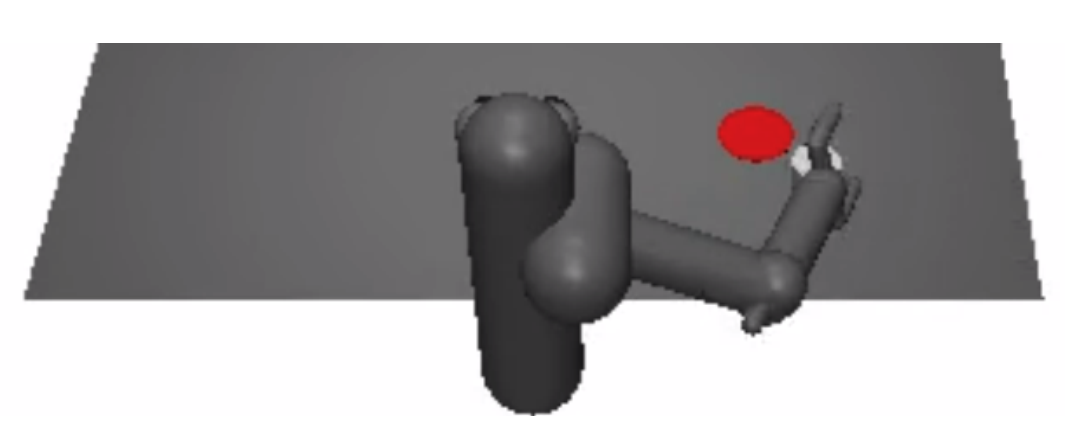}
    \includegraphics[height=0.115\textwidth]{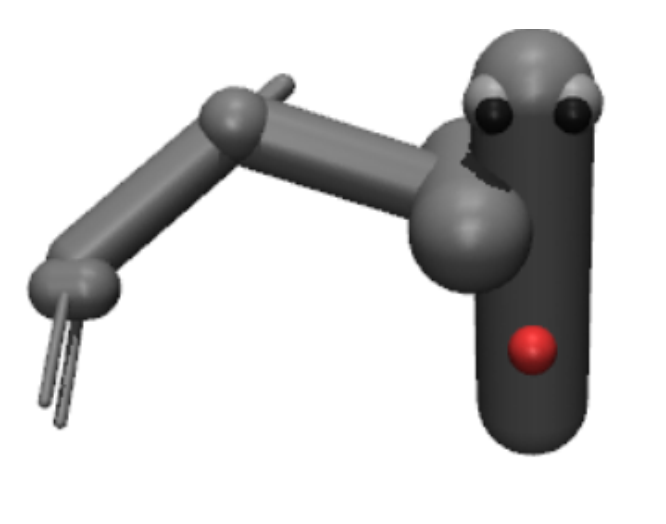}
    \includegraphics[height=0.115\textwidth,width=0.165\textwidth]{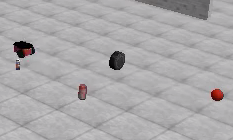}
    \includegraphics[height=0.115\textwidth,width=0.165\textwidth]{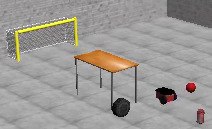}
    \includegraphics[height=0.115\textwidth]{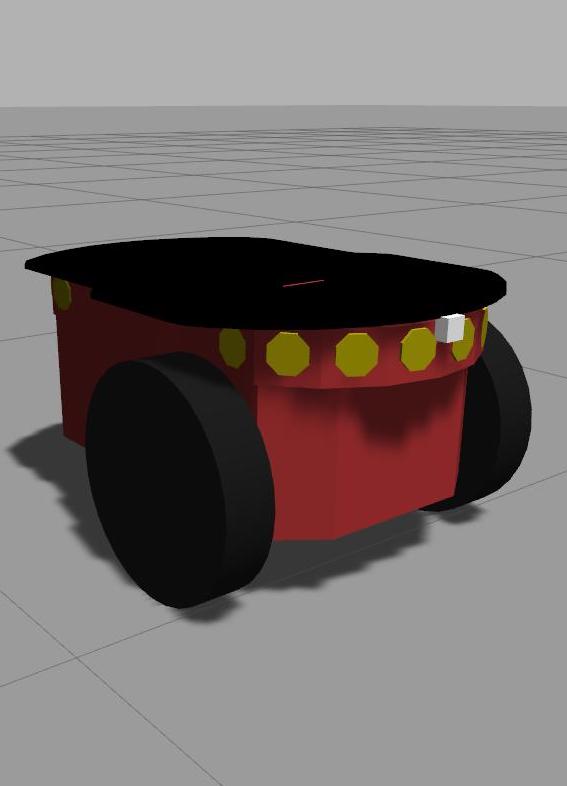}
    \caption{The four tasks we evaluate on. From left to right: \textsc{7-DOF
    Pusher}, \textsc{7-DOF Reacher}, \textsc{GoalTask}, and
    \textsc{KickBall}. The first two tasks simulate a one-armed \textsc{PR2} robot
    environment while the last two are in the \textsc{SocialRobot}
    environment. The final picture shows a closeup of the \textsc{Pioneer2dx}
    robot used in \textsc{SocialRobot}.}
    \label{fig:environment pictures}
\end{figure}

%\begin{wrapfigure}{R}{0.4\textwidth}
%    \centering
%    \includegraphics[height=0.15\textwidth]{images/environments/fetch.png}
%    \includegraphics[height=0.15\textwidth]{images/environments/pioneer2dx.jpg}
%    \caption{The two robots used in our tasks: 7-DoF PR2 arm and \texttt{Pioneer2dx}. Both have real-robot counterparts.}
%    \label{fig:robots}
%\end{wrapfigure}

\textbf{Environments}\ We evaluate success rate and sample efficiency across
two environment suites, as shown in Figure~\ref{fig:environment pictures}.
Important details are presented here with more information in appendix
Section \ref{sec:more environment details}.
The first suite consists of two 7-DOF reaching and pushing environments
evaluated in \citet{kurtl2018deep}. They both emulate a one-armed PR2 robot.
The tasks have sparse rewards: the agent gets a reward of 0 at every timestep
where the goal is not achieved, and 1 upon achieved. There is also a small
$L_2$ action penalty applied. In \textsc{7-DOF Reacher}, the goal is achieved
when the gripper reaches a 3D goal position. In \textsc{7-DOF Pusher}, the
goal is to push an object to a 3D goal position. Episodes have a fixed length of 100; 
a success of an episode is defined to be if the goal is achieved at the \emph{final} step of the
episode.

We also propose another suite of environments called \textsc{SocialRobot}~\footnote{\url{https://github.com/HorizonRobotics/SocialRobot}}. We
construct two sparse reward robotic navigation and manipulation tasks,
\textsc{GoalTask} and \textsc{KickBall}. In \textsc{GoalTask}, the agent gets
a reward of 1 when it successfully navigates to a goal, -1 if the goal
becomes too far, -0.5 every time it is too close to a distractor object, and
0 otherwise. In \textsc{KickBall}, the agent receives a reward of 1 for
successfully pushing a ball into the goal, 0 otherwise, and has the same
distractor object penalty. At the beginning of each episode, both the agent
and the ball are spawned randomly. Both environments contain a small $L_2$
action penalty, and terminate an episode upon a success.

%Fig.~\ref{fig:robots} shows the two robots used by the tasks. 
%Overall, according to the nature of action space, we divide the six tasks into two groups:

%\begin{compactenum}[a)]
%    \item 3D Euclidian action space: \texttt{FetchReach}, \texttt{FetchPush}, and \texttt{FetchSlide}.
%    \item General action space: \texttt{FetchPickAndPlace}, \texttt{GoalTask}, and \texttt{KickBall}.
%\end{compactenum}
%
%Potentially, the regular structure of a) can be more easily exploited by some simple trick such as action repetition, which we will include as baselines for comparison.
%

\textbf{Comparison methods}\ One baseline algorithm for comparison is
standard SAC \citep{haarnoja2018soft}, the building block of our hierarchical
method. To verify if our worker policy can just be replaced with a na\"ive
action repetition strategy, we compare with SAC+ActRepeat with an action
repetition for the same length $K$ as our option interval.
We also compare against HIRO \citep{nachum2018dataefficient}, a data
efficient hierarchical method with importance-based option relabeling, and
HiPPO \citep{Li2020} which trains the lower level and higher level policies
together with one unified PPO-based objective. Both are state-of-the-art
hierarchical methods proposed to solve sparse reward tasks. Similar to our
work, HiPPO makes no assumptions about options, however it utilizes a
discrete option space and its options are trained with environment reward.

\iffalse
\haonan{We could just leave this paragraph out. HIRO actually makes quite
some assumptions about the state/goal space.} We compare against HIRO and
HiPPO specifically because they make few assumptions about the task (just
like our method) relative to other hierarchical learning works, and they
perform well on their respective sparse reward tasks against other baseline
methods.
\fi 

%We note that our approach is complementary to
%Hindsight Experience Replay (HER) \cite{andrychowicz2017hindsight}, which
%assumes that the agent has access to the reward function, and that there is a
%mapping from any state $s$ to a satsified goal. In order to keep \method\ as
%general as possible for any sparse reward task, we do not make these
%assumptions. However given those assumptions, \method\ can be combined with
%HER.

%For all methods compared, data is gathered by 20 parallel actors in the
%\textsc{7-DOF} environments, and 10 parallel actors in the
%\textsc{SocialRobot} environments.
We implement HIDIO based on an RL framework called ALF~\footnote{\url{https://github.com/HorizonRobotics/alf}}.
A comprehensive hyperparameter search is
performed for every method, with a far greater search space over HiPPO and
HIRO than our method \method\ to ensure maximum fairness in comparison; details are
presented in Appendix~\ref{sec:hyperparameters}.

\textbf{Evaluation}\ \ For every evaluation point during training, we
evaluate the agent with current deterministic policies (by taking $\argmax$
of action distributions) for a fixed number of episodes and compute the mean
success rate. We plot the mean evaluation curve over 3 randomly seeded runs
with standard deviations shown as the shaded area around the curve.

\iffalse
We now ablate out various components of \method\ to determine their
importance in improving sample efficiency and success rate in our
environments.
\fi

\subsection{Worker Design Choices}
\label{sec:worker_ablations}
We ask and answer questions about the design choices in \method\ specific to
the worker policy \pilow.

\begin{figure}
    \centering
    \includegraphics[width=0.24\textwidth]{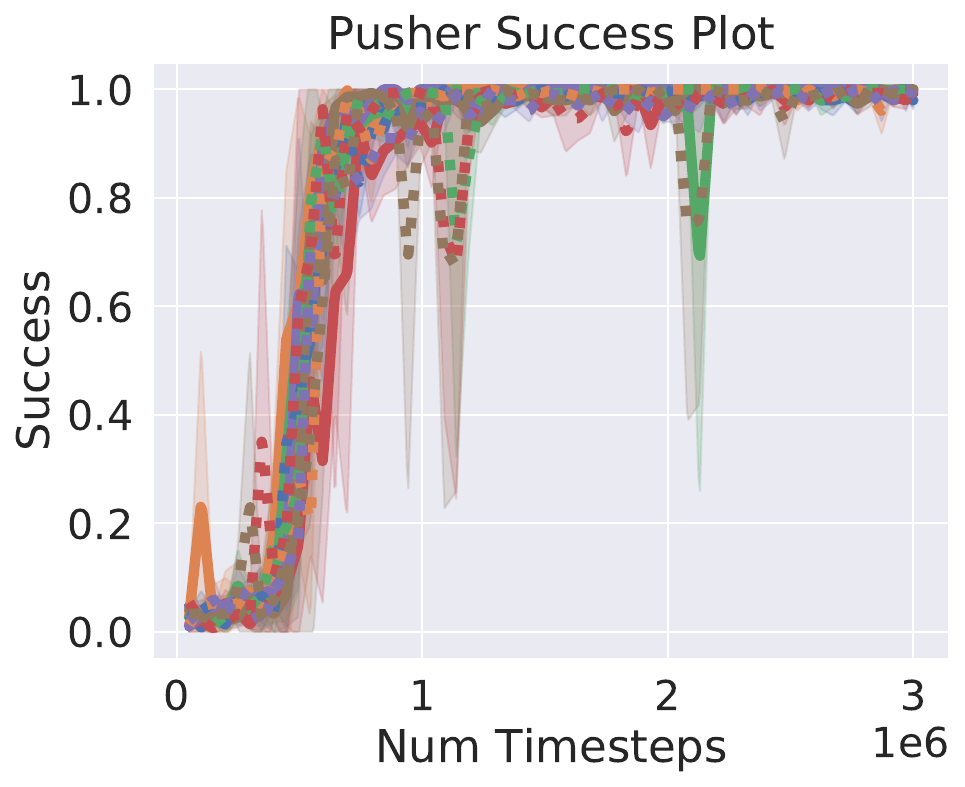}
    \includegraphics[width=0.24\textwidth]{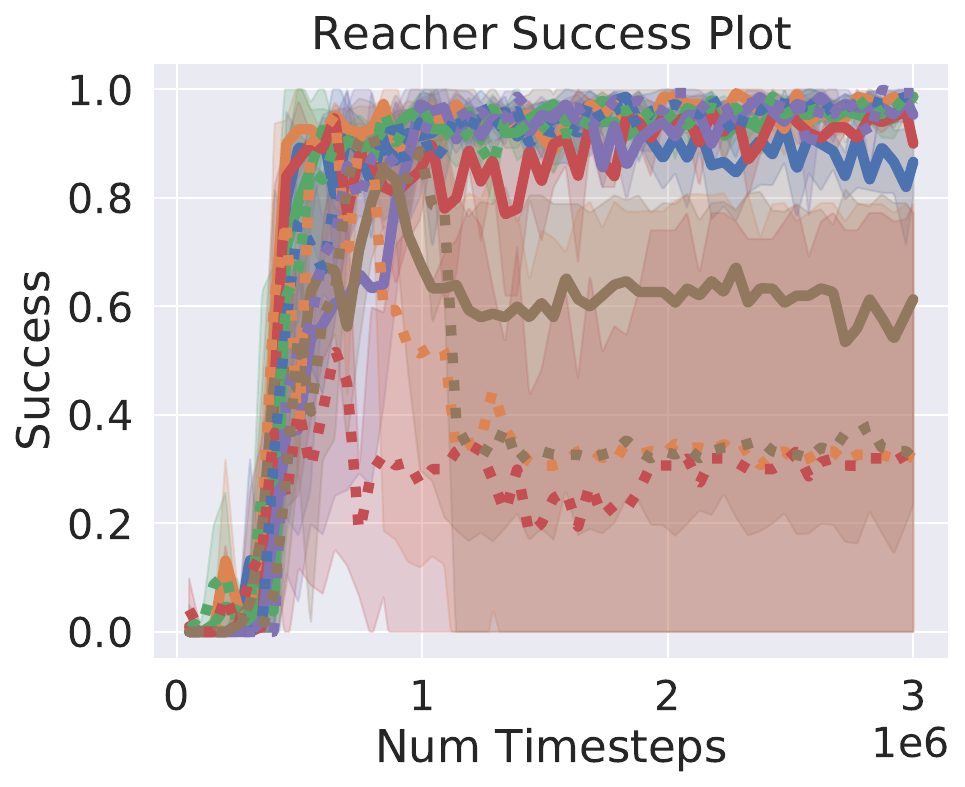}
    \includegraphics[width=0.24\textwidth]{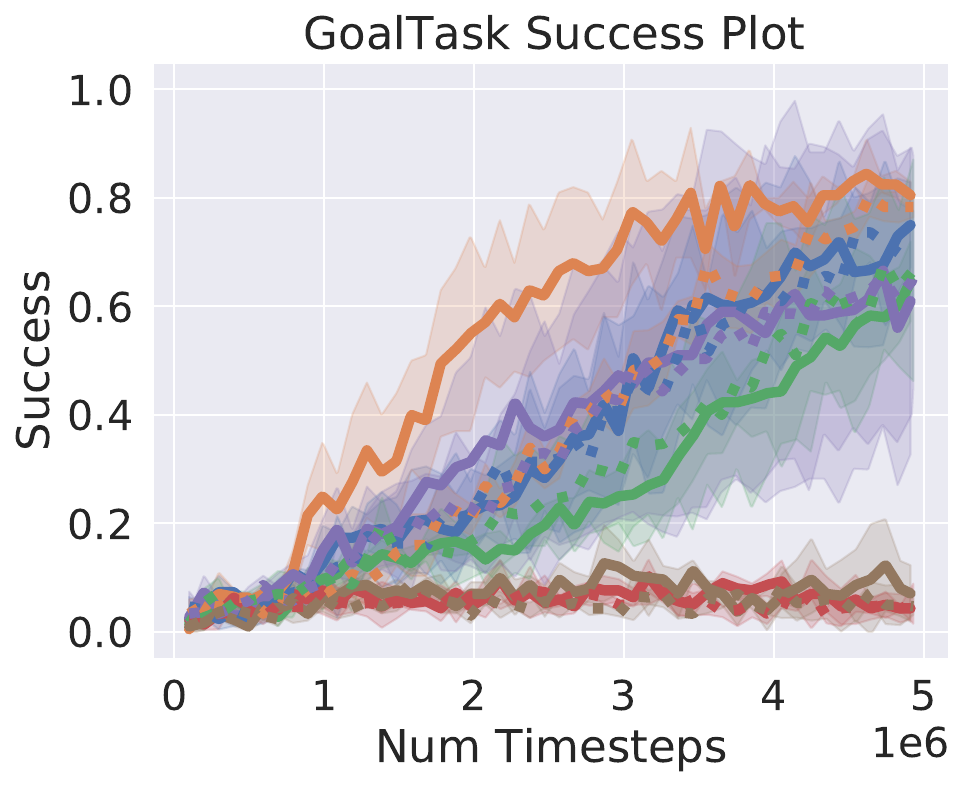}
    \includegraphics[width=0.24\textwidth]{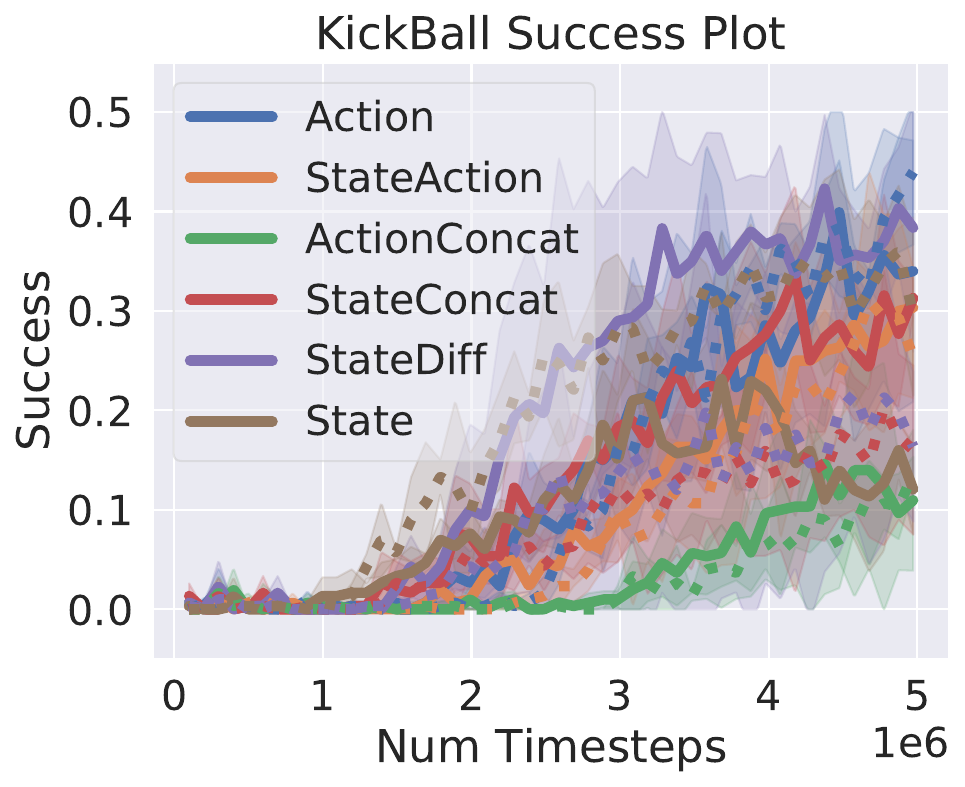}
    \caption{Comparison of all discriminator features against each
    other across the four environments. Solid lines indicate hard
    short-sighted workers (\texttt{Hard}), dotted lines indicated soft
    short-sighted workers (\texttt{Soft}).}
    \label{fig:feature comparison}
\end{figure}

1. \textit{What sub-trajectory feature results in good option discovery?} We
evaluate all six features proposed in Section~\ref{sec:discriminator} in all four
environments.
These features are selected to evaluate how different types of subtrajectory
information affect option discovery and final performance. They
encompass varying types of both local and global subtrajectory information.
%Other work has utilized full observation trajectory information for skill discrimination
%\citep{warde-farley2018, achiam2018variational}, of which our
%feature extractor \texttt{StateConcat} is most similar to. 
We plot comparisons of sample efficiency and final performance in
Figure~\ref{fig:feature comparison} across all environments (solid lines),
finding that \texttt{Action}, \texttt{StateAction}, and \texttt{StateDiff}
are generally among the top performers. \texttt{StateAction} includes the
current action and next state, encouraging \pilow\ to differentiate its
options with different actions even at similar states. Similarly,
\texttt{Action} includes the option initial state and current action,
encouraging option diversity by differentiating between actions
conditioned on initial states. Meanwhile \texttt{StateDiff} simply encodes the
difference between the next and current state, encouraging \pilow\ to produce
options with different state changes at each step.

2. \textit{How do soft shortsighted workers (\texttt{Soft}) compare against
hard shortsighted workers (\texttt{Hard})?} In Figure~\ref{fig:feature
comparison}, we plot all features with \texttt{Soft} in dotted lines. We can
see that in general there is not much difference in performance between
\texttt{Hard} and \texttt{Soft} except some extra instability of
\texttt{Soft} in \textsc{Reacher} regarding the \texttt{StateConcat} and
\texttt{State} features. One reason of this similar general performance could
be that since our options are very short-term in \texttt{Hard}, the
scheduler policy has the opportunity of switching to a good option before the
current one leads to bad consequences. In a few cases, \texttt{Hard} seems
better learned, perhaps due to an easier value bootstrapping for the worker.

\begin{figure}
    \centering
    \includegraphics[width=0.24\textwidth]{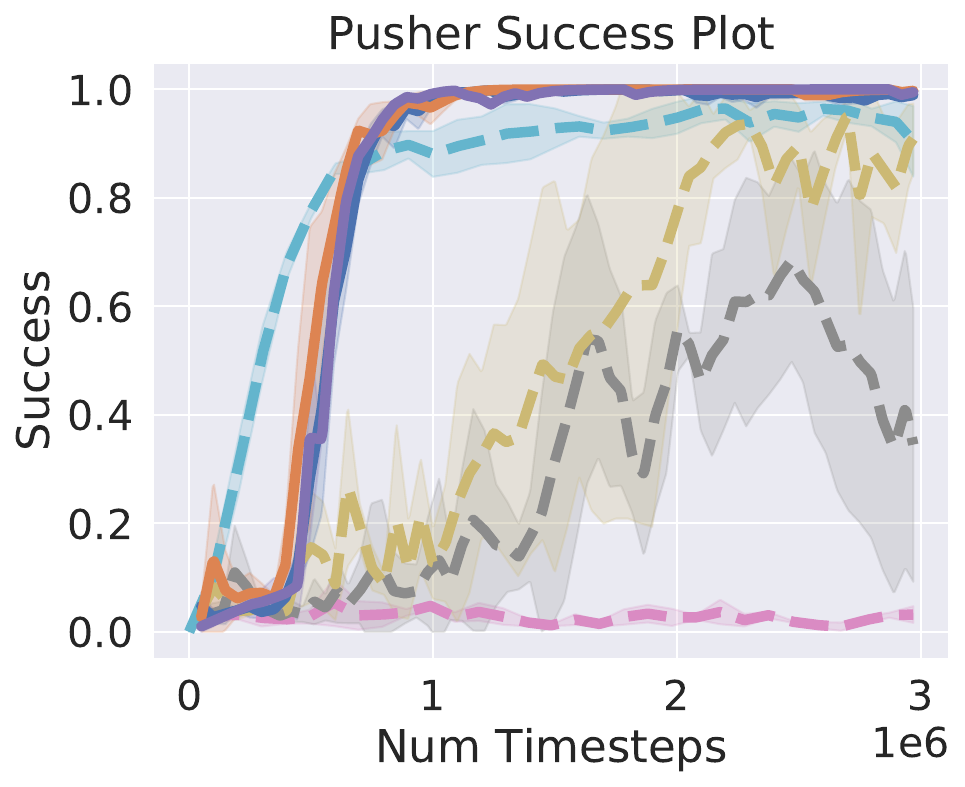}
    \includegraphics[width=0.24\textwidth]{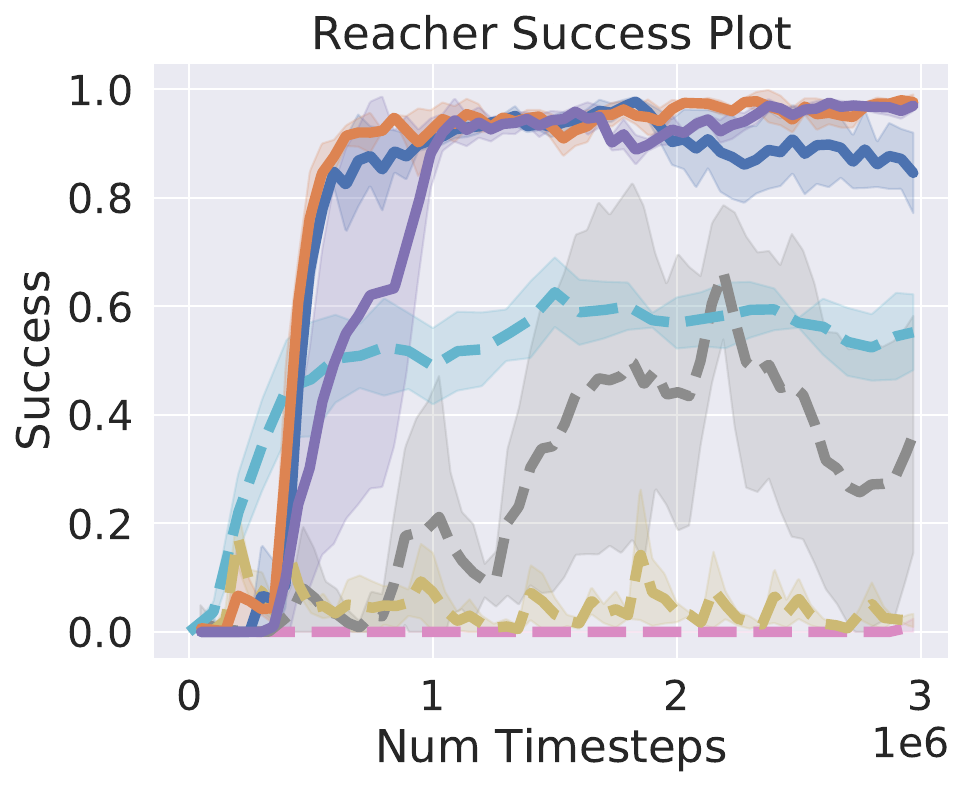}
    \includegraphics[width=0.24\textwidth]{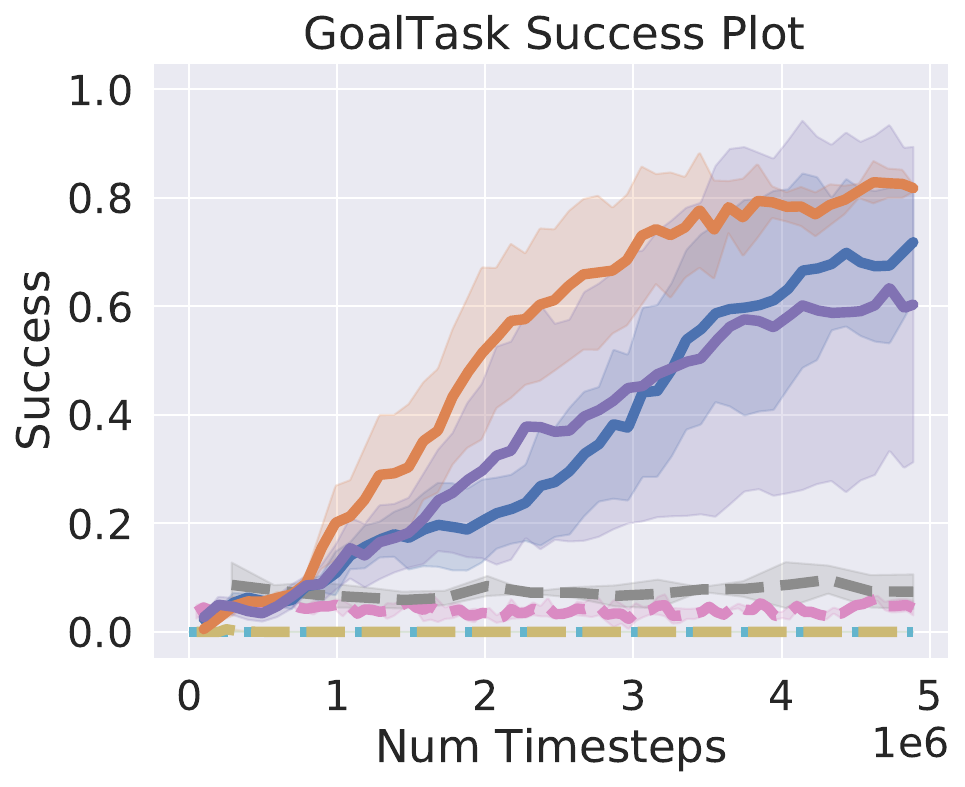}
    \includegraphics[width=0.24\textwidth]{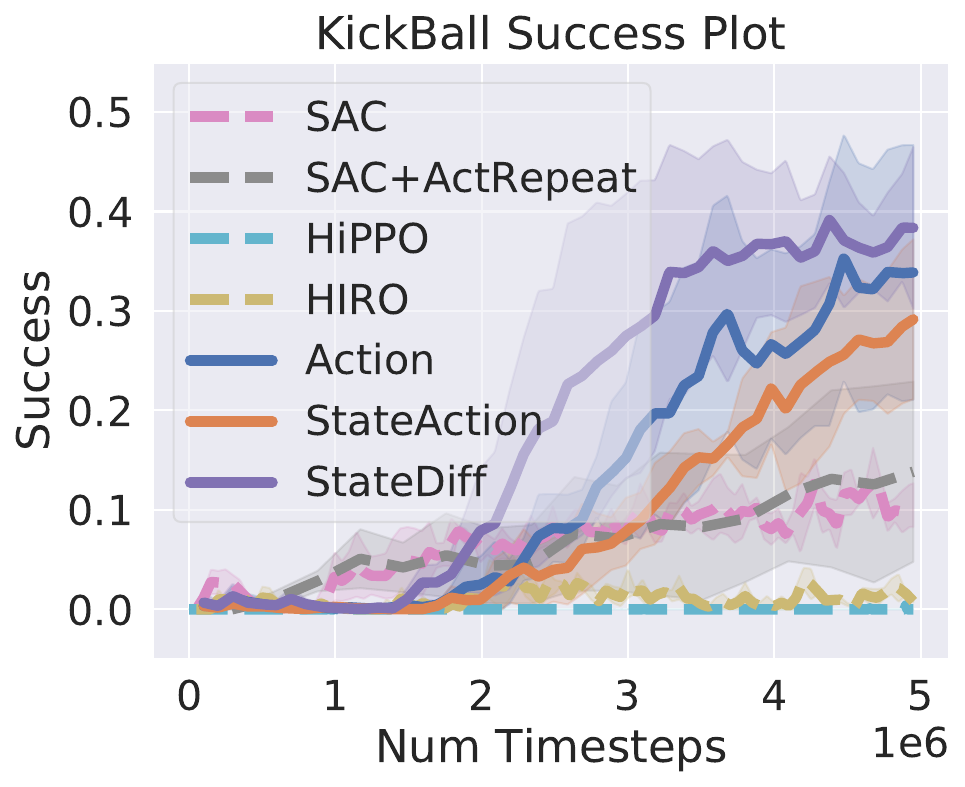}
    \caption{Comparisons of the mean success rates of three features of
    \framework\ (\texttt{Action}, \texttt{StateAction}, \texttt{StateDiff};
    solid lines) against other methods (dashed lines).}
    \label{fig:all comparison}
    \vspace{-8pt}
\end{figure}
\subsection{Comparison results} 
We compare our three best sub-trajectory features of \texttt{Hard},
in Section~\ref{sec:worker_ablations}, against the
SAC baselines and hierarchical RL methods across all four environments in
Figure~\ref{fig:all comparison}. Generally we see that \method\ (solid lines)
achieves greater final performance with superior sample efficiency than the
compared methods. Both SAC and SAC+ActRepeat perform poorly across all
environments, and all baseline methods perform significantly worse than
\method\ on \textsc{Reacher}, \textsc{GoalTask}, and \textsc{KickBall}.

In \textsc{Pusher}, HiPPO displays competitive performance, rapidly improving
from the start. However, all three \method\ instantiations achieve nearly
100\% success rates while HiPPO is unable to do so. Furthermore, HIRO and
SAC+ActRepeat take much longer to start performing well, but never achieve
similar success rates as \method.
\method\ is able to solve \textsc{Reacher} while HiPPO achieves only about a
60\% success rate at best. Meanwhile, HIRO, SAC+ActRepeat, and SAC are
unstable or non-competitive. \textsc{Reacher} is a difficult exploration
problem as the arm starts far from the goal position, and we see that
\method's automatically discovered options ease exploration for the higher
level policy to consistently reach the goal. % Finally, all three
\method\ performs well on \textsc{GoalTask}, achieving 60-80\%
success rates, while the task is too challenging for every other method. In
\textsc{KickBall}, the most challenging task, \method\ achieves
30-40\% success rates while every other learns poorly again,
highlighting the need for the intrinsic option discovery of \method\ in these
environments.

In summary, \method\ demonstrates greater sample efficiency and final reward
gains over all other baseline methods. Regular RL (SAC) fails on all four
environments, and while HiPPO is a strong baseline on \textsc{Pusher} and
\textsc{Reacher}, it is still outperformed in both by \method. All other
methods fail on \textsc{GoalTask} and \textsc{KickBall}, while \method\ is
able to learn and perform better in both. This demonstrates the importance of
the intrinsic, short-term option discovery employed by \method, where the
options are diverse enough to be useful for both exploration and task
completion.

\begin{figure}
    \centering
    \includegraphics[width=1.0\textwidth]{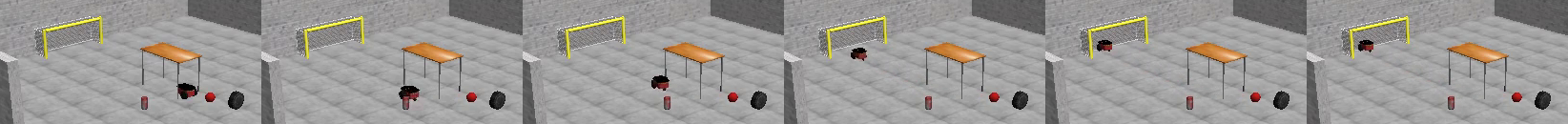}
    \includegraphics[width=1.0\textwidth]{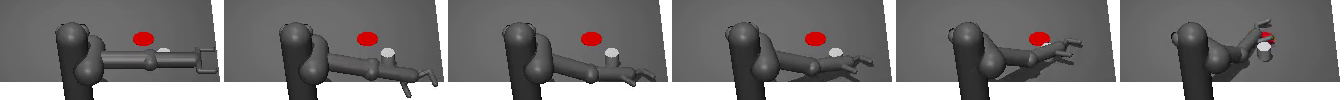}
    \caption{Two example options from the \texttt{StateAction}
    instantiation on \textsc{KickBall} (top) and \textsc{Pusher}
    (bottom). The top option navigates directly to the goal by bypassing
    obstructions along the way and the bottom option sweeps the puck towards
    one direction.}
    \label{fig:skill visualization}
    \vspace{-8pt}
\end{figure}

\begin{wrapfigure}{R}{0.44\textwidth}
    \centering
    \includegraphics[width=0.21\textwidth]{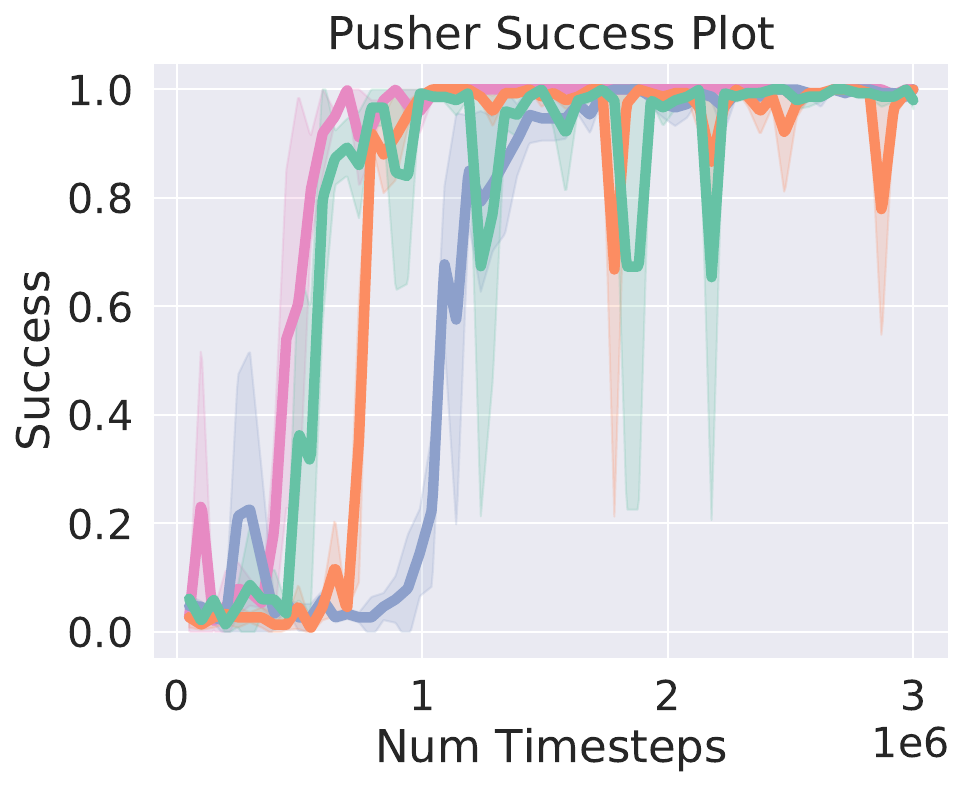}
    \includegraphics[width=0.21\textwidth]{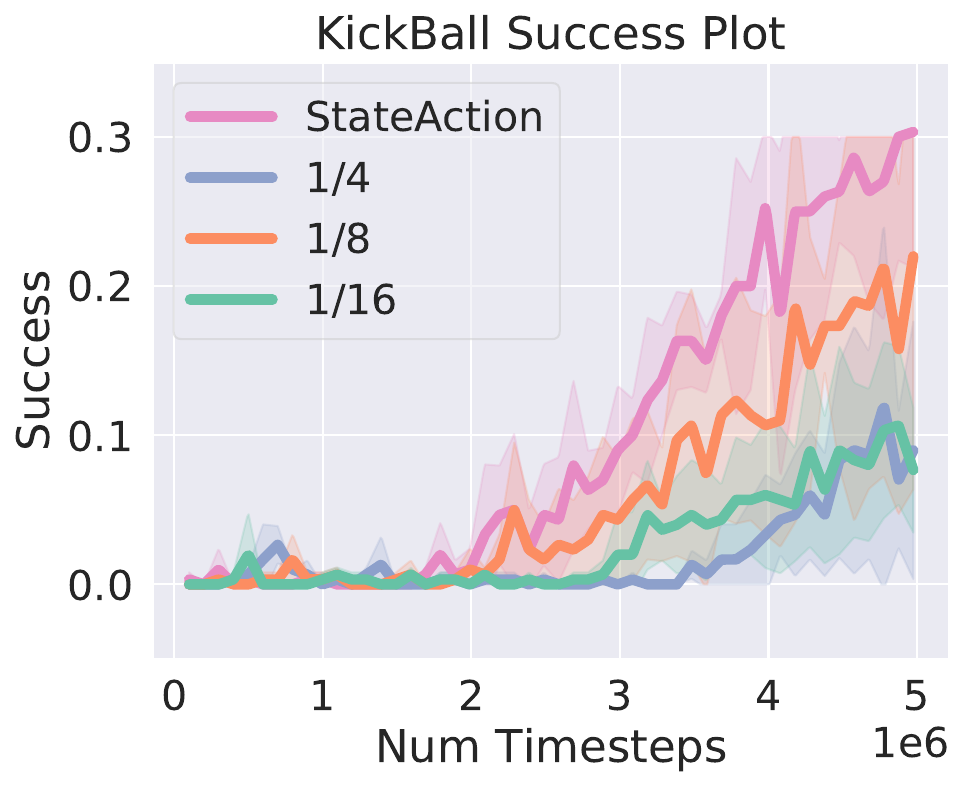}
    \caption{Pretraining baseline comparison at fractions
    $\{\frac{1}{16},\frac{1}{8},\frac{1}{4}\}$ of the total number of
    training time steps.}
    \label{fig:pretraining}
    \vspace{-8pt}
\end{wrapfigure}

\subsection{Joint \pilow\ and \pihi\ training} 
We ask the next question:
\textit{is jointly training $\pihi$ and $\pilow$ necessary?} To answer this,
we compare \method\ against a pre-training baseline where we first pre-train
\pilow, with uniformly sampled options $\bfu$ for a portion $\rho$ of total
numbers of training time steps, and then fix \pilow\ while training \pihi\ for the
remaining $(1-\rho)$ time steps. This is essentially using pre-trained options
for downstream higher-level tasks as demonstrated in DIAYN
\citep{eysenbach2018diversity}. We conduct this experiment with the
\texttt{StateAction} feature on both \textsc{KickBall} and
\textsc{Pusher}, with $\rho=\{\frac{1}{16},\frac{1}{8},\frac{1}{4}\}$. The
results are shown in Figure~\ref{fig:pretraining}. We can see that in
\textsc{Pusher}, fewer pre-training time steps are more sample efficient, as
the environment is simple and options can be learned from a small amount of
samples. The nature of \textsc{Pusher} also only requires 
options that can be learned independent of the scheduler policy evolution.
Nevertheless, the pre-training baselines seem less stable. In
\textsc{KickBall}, the optimal pre-training baseline is on $\rho=\frac{1}{8}$
of the total time steps. However without the joint training scheme of
\method, the learned options are unable to be used as efficiently for the
difficult obstacle avoidance, navigation, and ball manipulation subtasks 
required for performing well.

\subsection{Option Behaviors} 
\label{sec:skill visualizations}
Finally, since options discovered by \method\ in our sparse reward
environments help it achieve superior performance, we ask, \textit{what do
useful options look like?} To answer this question, after training, we sample options from 
the scheduler \pihi\ to visualize their behaviors in different environments
in Figure~\ref{fig:skill visualization}. For each sampled option $\bfu$, we
fix it until the end of an episode and use the worker $\pilow$ to output
actions given $\bfu$. We can see that the options learned by \method\ are
low-level navigation and manipulation skills useful for the respective
environments. We present more visualizations in Figure~\ref{fig:extra skill visualization} and
more analysis in Section~\ref{sec:extra skill visualizations} in the appendix. Furthermore, 
we present an analysis of task performance for 
different option lengths in appendix Section~\ref{sec:option length ablation} and Figures~\ref{fig:option length ablation} and~\ref{fig:option length trajectory vis}.
%===============================================================================
%!TEX root=main.tex
\section{Related Work}
\label{sec:related work} 
\paragraph{Hierarchical RL} Much of the previous work in hRL
makes assumptions about the task structure and/or the skills needed to solve the task.
%\citep{florensa2017stochastic, Riedmiller2018, lee2018composing,
%hausman2018learning,Lee2020Learning,sohn2018hierarchical,ghavamzadeh2003hierarchical}.
While obtaining promising results under specific settings, they may have
difficulties with different scenarios. 
For example, SAC-X \citep{Riedmiller2018} requires manually designing
auxiliary subtasks as skills to solve a given downstream task. SNN4HRL
\citep{florensa2017stochastic} is geared towards tasks with pre-training and 
downstream components. \citet{lee2018composing,Lee2020Learning} learns to
modulate or compose given primitive skills that are customized for their
particular robotics tasks. 
%\citet{hausman2018learning}
%assumes the access to an initial set of tasks with accompanying reward
%functions so as to learn to interpolate and/or sequence those tasks to create
%new ones. 
\citet{ghavamzadeh2003hierarchical} and
\citet{sohn2018hierarchical} operate under the assumption that tasks can be
manually decomposed into subtasks.

%Other works assume access to a set of demonstrations
%\citep{sharma2018directedinfo, le2018hierarchical, merel2018hierarchical,
%ranchod2015nonparametric} which can lead to very good task performance, but
%demonstrations can be very difficult to obtain.

The feudal reinforcement learning proposal~\citep{Dayan1993} has inspired
another line of works \citep{Vezhnevets2017,nachum2018dataefficient,Levy2019,rafati2019learning}
which make higher-level manager policies output goals for lower-level worker
policies to achieve. Usually the goal space is a subspace of the state space
or defined according to the task so that lower-level rewards are easy to compute. 
This requirement of manually
``grounding'' goals in the environment poses generalization challenges for
tasks that cannot be decomposed into state or goal-reaching.

The MAXQ decomposition \citep{dietterich2000hierarchical} defines 
an hRL task decomposition by breaking up the target MDP into a hierarchy
of smaller MDPs such that the value function in the target MDP is represented
as the sum of the value functions of the smaller ones. This has inspired 
works that use such decompositions \citep{mehta2008automatic, winder2020planning, li2017efficient}
to learn structured, hierarchical world models or policies to complete target tasks or 
perform transfer learning. 
However, building such hierarchies makes these methods limited to MDPs with discrete action
spaces.

Our method \framework\ makes few assumptions about the specific task
at hand. It follows from the options framework \citep{Sutton1999}, which has
recently been applied to continuous domains \citep{bacon2016optioncritic},
spawning a diverse set of recent hierarchical options methods
\citep{bagaria2019option, klissarov2017learnings, riemer2018learning,
tiwari2018natural, jain2018safe}. \framework\ automatically learns
intrinsic options that avoids having explicit initiation or termination
policies dependent on the task at hand. HiPPO \citep{Li2020}, like
\framework, also makes no major assumptions about the task, but does not employ 
self-supervised learning for training the lower-level policy.

\paragraph{Self-supervised option/skill discovery} There are also plenty of prior works
which attempt to learn skills or options without task reward.
%\citep{eysenbach2018diversity,Gregor2016,achiam2018variational, Sharma2019, warde-farley2018}.
DIAYN \citep{eysenbach2018diversity} and VIC \citep{Gregor2016} learn skills by maximizing
the mutual information between trajectory states and their corresponding skills. 
VALOR \citep{achiam2018variational} learns options by maximizing the
probability of options given their resulting observation
trajectory. DADS \citep{Sharma2019} learns skills that are predictable by
dynamics models. DISCERN \citep{warde-farley2018} maximizes the mutual information between goal 
and option termination states to learn a goal-conditioned reward function. 
\citet{pmlr-v32-brunskill14} learns options in discrete
MDPs that are guaranteed to improve a measure of sample complexity. Portable Option Discovery \citep{Topin2015PortableOD} discovers options by merging options from 
source policies to apply to some target domain.
\cite{eysenbach2018diversity,
achiam2018variational,Sharma2019,lynch2020learning} demonstrate pre-trained options to be
useful for hRL. These methods usually pre-train options in an initial stage
separate from downstream task learning; few works directly
integrate option discovery into a hierarchical setting.
For higher dimensional input domains, \citet{lynch2020learning} learns 
options from human-collected robot interaction data for image-based, goal-conditioned tasks, and  
\citet{chuck2020hypothesisdriven} learns a hierarchy of options 
by discovering objects from environment images and forming options which can manipulate them. \method can also be applied to image-based environments by replacing fully-connected layers with convolutional layers in the early stages of the policy and discriminator networks. However, we leave this to future work to address possible practical challenges arising in this process. 

%Finally, there have been recent works that aim to learn these self-supervised 
%skill embeddings specifically for hRL. 
%TAIC \citep{wang2020learning} learns an embedding from action sequences by
%maximizing the mutual information between options and the first and final
%state of the option trajectory. 
%SeCTAR \citep{coreyes2018selfconsistent}
%learns a latent representation of observation trajectories for model-based,
%hierarchical planning. %\citet{hausman2018learning, lynch2020learning} learns
%an observation-based latent embedding for multi-task hierarchical
%reinforcement learning. 

%===============================================================================

\section{Conclusion}
\label{sec:conclusion}
Towards solving difficult sparse reward tasks, we propose a new 
hierarchical reinforcement learning method,
\method, which can learn task-agnostic options in a self-supervised manner
and simultaneously learn to utilize them to solve tasks. We evaluate several different
instantiations of the discriminator of \method\ for providing intrinsic rewards for training
the lower-level worker policy. We demonstrate the effectiveness of \method\ compared against 
other reinforcement learning methods in achieving high rewards with 
better sample efficiency across a variety of robotic navigation and 
manipulation tasks.

%\section*{Acknowledgements}
%We would like to thank Ofir Nachum and Jonas Gehring for assisting us with running HIRO, Alex Li for helping with the HiPPO implementation, and  Haichao Zhang, Qinxun Bai, and Le Zhao for valuable discussions during this project.
%===============================================================================

% The maximum paper length is 8 pages excluding references and acknowledgements, and 10 pages including references and acknowledgements

\clearpage
% The acknowledgments are automatically included only in the final version of the paper.

%===============================================================================

% no \bibliographystyle is required, since the corl style is automatically used.
\bibliography{references}
\bibliographystyle{iclr2021_conference}
\appendix
%!TEX root=main.tex
\section{Pseudo code for \framework}
\label{sec:pseudo_code}
%\begin{wrapfigure}{R}{0.55\textwidth}
%    \begin{minipage}{0.55\textwidth}
    \begin{algorithm}[H]
        \SetAlgoLined
        \textbf{Input:}\\
        \resizebox{\textwidth}{!}{
        \begin{tabular}{llllll}
            $T$ & Episode length & $M$ & Batches per iteration & $\pilow(\bfa_{h,k}|\mbfs_{h,k},\mbfa_{h, k-1}, \bfu_h)$ & Worker\\
            $B$ & Batch size & $\alpha$ & Learning rate & $q_{\psi}(\bfu_h|\mbfa_{h,k},\mbfs_{h,k+1})$ & Discriminator\\
            $K$ & Option interval  & $\mathcal{P}(\bfs_{h,k+1}|\bfs_{s,k},\bfa_{h,k})$ & Environment dynamics & $\pihi(\bfu_h|\bfs_{h,0})$ & Scheduler\\
        \end{tabular}
        }
        \\
        \textbf{Output:} Learned parameters $\theta$, $\phi$, and $\psi$.\\
        \textbf{Initialize:} Random model parameters $\theta$, $\phi$, and $\psi$; empty replay buffers $\mathcal{D}_{\text{scheduler}}$ and  $\mathcal{D}_{\text{worker}}$.\\
        \While{termination not met}{
            \tcc{Data collection}
            \For{scheduler step $h=0..\frac{T}{K}-1$}{
                Sample an option $\bfu_h \sim \pihi(\cdot|\bfs_{h,0})$.\\
                \For{worker step $k=0..K-1$}{
                    Sample an action $\bfa_{h,k}\sim \pilow(\cdot|\mbfs_{h,k}, \mbfa_{h, k-1}, \bfu_h)$.\\
                    Step through the environment $\bfs_{h,k+1}\sim \mathcal{P}(\cdot|\bfs_{h,k},\bfa_{h,k})$.\\
                    $\mbfa_{h,k},\mbfs_{h,k+1} \leftarrow [\mbfa_{h,k-1},\bfa_{h,k}], [\mbfs_{h,k},\bfs_{h,k+1}]$\\    
                    $\mathcal{D}_{\text{worker}} \leftarrow \mathcal{D}_{\text{worker}} \cup (\bfu_h,\mbfa_{h,k},\mbfs_{h,k+1})$\\
                }
                $R_h \leftarrow \sum_{k=0}^{K-1}r(\bfs_{h,k},\bfa_{h,k},\bfs_{h,k+1})$\\
                $\mathcal{D}_{\text{scheduler}} \leftarrow \mathcal{D}_{\text{scheduler}} \cup (\bfs_{h,0},\bfu_{h},\bfs_{h+1,0},R_h)$\\
            }
            \tcc{Model training}
            \For{batch $m=0..M-1$}{
                \tcc{Scheduler training}
                Uniformly sample transitions $\{(\bfs_t,\bfu_t,\bfs_{t+1})\}_{b=1}^B\sim \mathcal{D}_{\text{scheduler}}$.\\
                Compute gradient $\Delta\theta$ according to Eq.~\ref{eq:scheduler_objective}.\\
                Update models $\theta \leftarrow \theta + \alpha \Delta\theta$.\\
                \tcc{Worker training}
                Uniformly sample transitions $\{(\bfu_h,\mbfa_{h,k},\mbfs_{h,k+1})\}_{b=1}^B \sim \mathcal{D}_{\text{worker}}$.\\
                Compute intrinsic rewards $r^{lo}_{h,k}\leftarrow q_{\psi}(\bfu_h|\mbfa_{h,k},\mbfs_{h,k+1})$.\\
                Compute gradient $\Delta\psi$ and $\Delta\phi$ according to Eq.~\ref{eq:worker_objective}.\\
                Update models $\phi\leftarrow \phi+\alpha\Delta\phi$ and $\psi\leftarrow \psi+\alpha\Delta\psi$.\\
            }
        }
        \caption{Hierarchical RL with Intrinsic Options Discovery}
    \end{algorithm}
%    \end{minipage}
%\end{wrapfigure}

\section{More environment details} \label{sec:more environment details}
\subsubsection{Pusher and Reacher}
These environments both have a time horizon of 100 with no early termination:
each episode always runs for 100 steps regardless of goal achievement. 
For both, a success is when the agent achieves the goal at the \emph{final} step of an episode. 
In \textsc{Reacher}, observations are 17-dimensional, including the positions, angles, and velocities of the robot arm, and in \textsc{Pusher} observations also include the 3D object position. Both include the goal position in the observation space. Actions are 7-dimensional vectors for joint velocity control. The action range is $[-20, 20]$ in \textsc{Reacher} 
and $[-2, 2]$ in \textsc{Pusher}.

There is an action penalty in both environments: at every timestep the squared 
$L_2$ norm of the agent action is subtracted from the reward. In \textsc{Pusher},
this penalty is multiplied by a coefficient of $0.001$. In \textsc{Reacher}, 
it's multiplied by $0.0001$. 

\subsubsection{GoalTask and KickBall}
For both \textsc{SocialRobot} environments, an episode terminates early when either a success is reached or the goal is out of range.
For each episode, the positions of all objects (including the agent) are randomly picked. Observations are 18-dimensional. In \textsc{GoalTask}, these observations include ego-centric positions, distances, and directions from the agent to different objects while in \textsc{KickBall}, they are absolute positions and directions.
In \textsc{KickBall}, the agent receives a reward of 1
for successfully pushing a ball into the goal (episode termination) and 0 otherwise. At the beginning of each 
episode, the ball is spawned randomly inside the neighborhood of the agent. Three distractor objects are included on the ground to increase task difficulty. In \textsc{GoalTask}, the number of distractor objects increases to 5.
Both environments contain a small $L_2$ action penalty: at every time step the squared $L_2$ norm of the agent action, multiplied by $0.01$, is subtracted from the reward.
\textsc{GoalTask} has a time horizon of
100 steps, while \textsc{KickBall}'s horizon is 200. Observations are 30-dimensional, including absolute poses and velocities of the goal, the ball, and the agent. Both \textsc{GoalTask} and \textsc{KickBall} use the same navigation robot \textsc{Pioneer2dx} which has 2-dimensional actions that control the angular velocities (scaled to $[-1, 1]$) of the two wheels. 

\section{Option Details}
\subsection{Option Length Ablation}
\label{sec:option length ablation}
\begin{figure}
    \centering
    \includegraphics[width=0.24\textwidth]{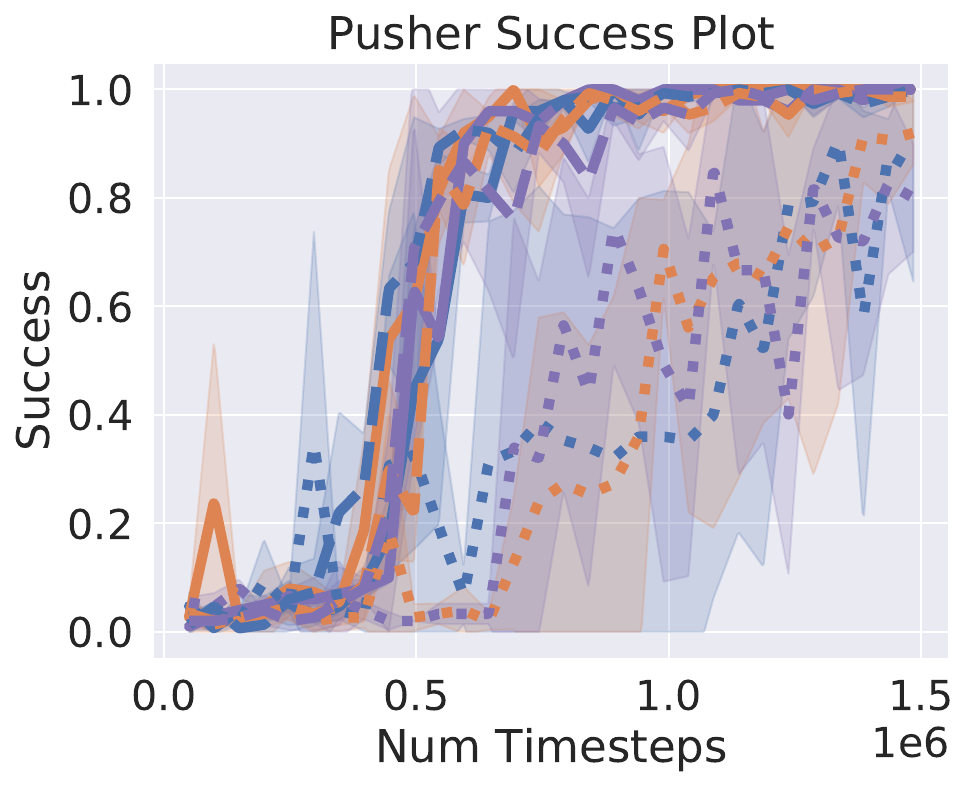}
    \includegraphics[width=0.24\textwidth]{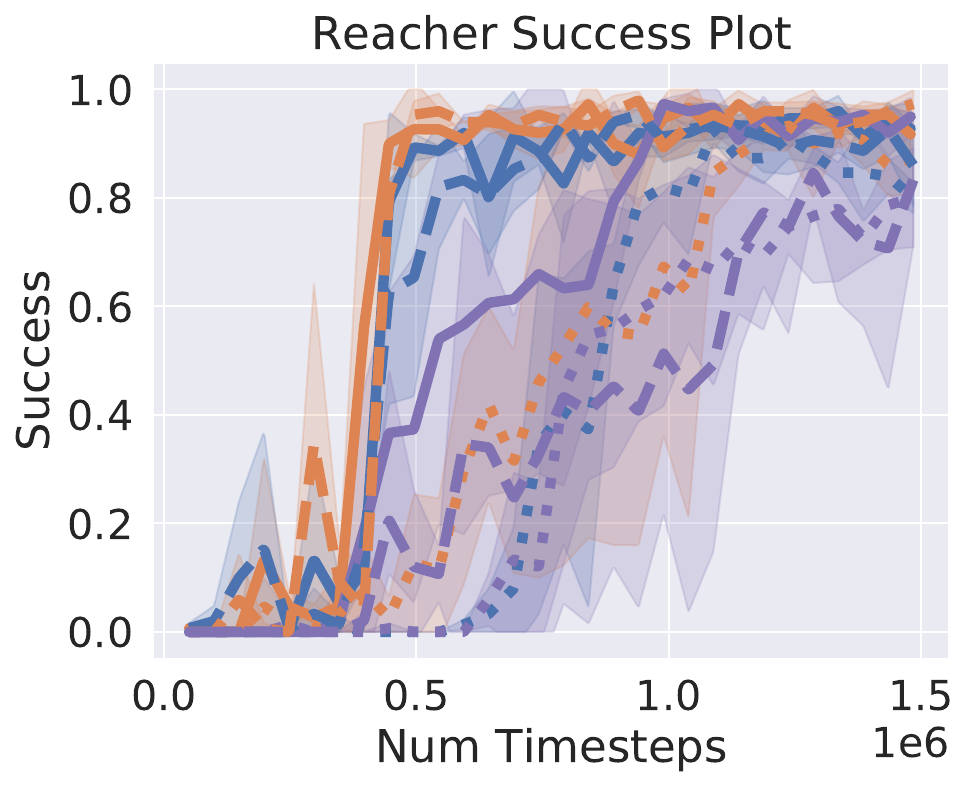}
    \includegraphics[width=0.24\textwidth]{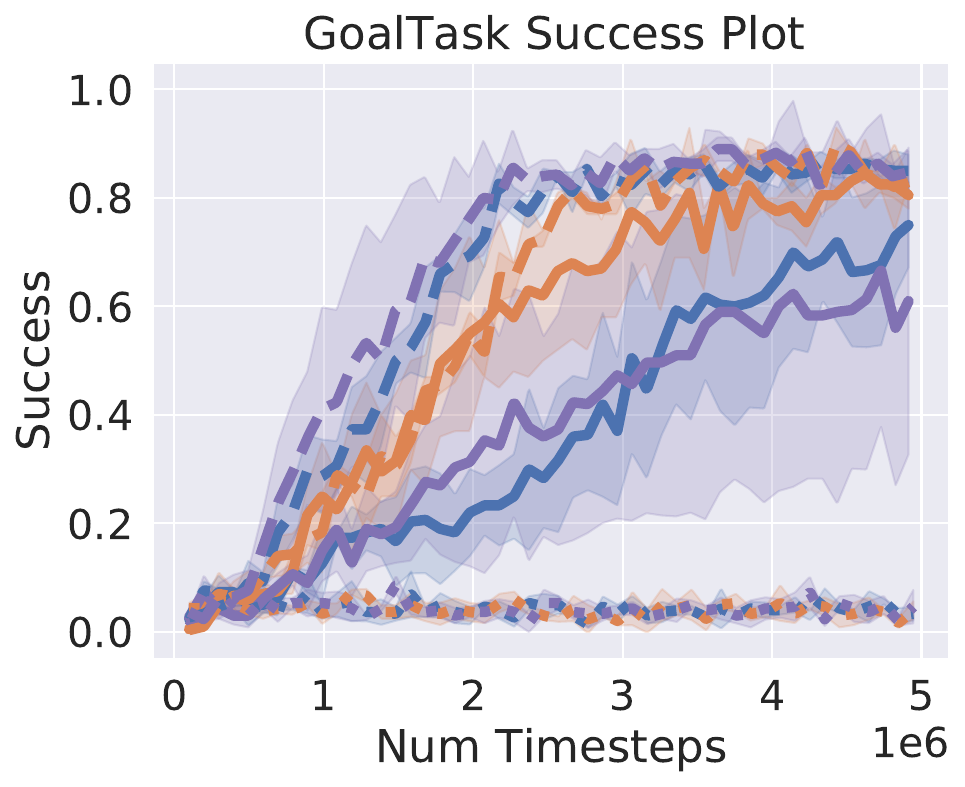}
    \includegraphics[width=0.24\textwidth]{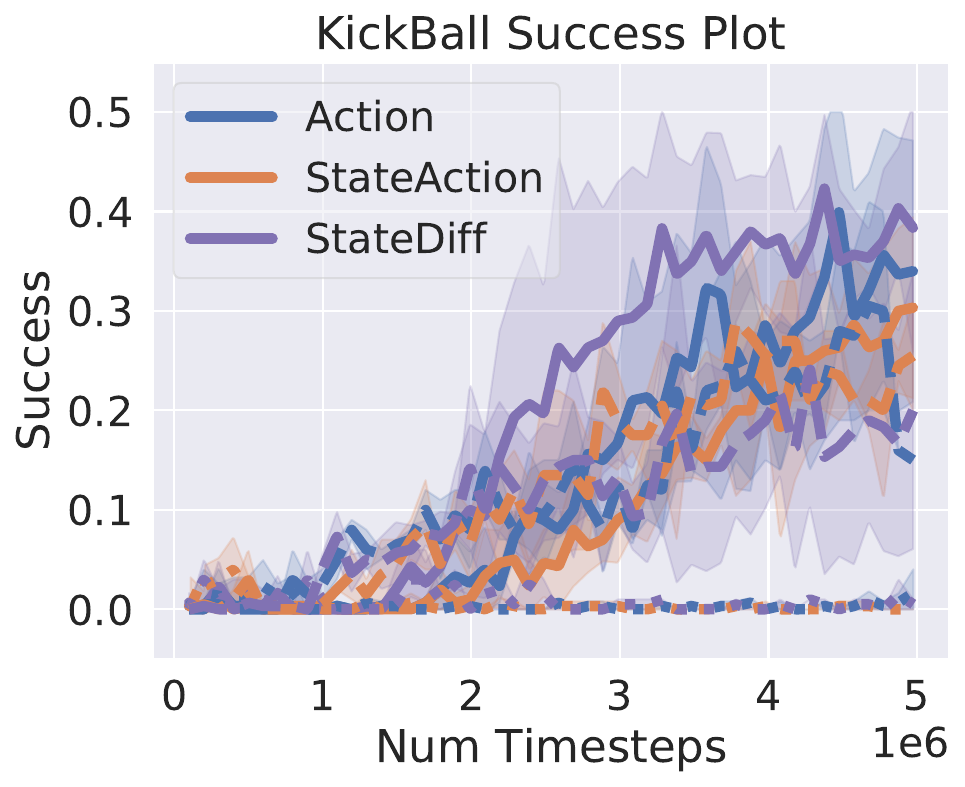}
    \caption{Comparisons of the mean success rates of three features of
    \framework\ (\texttt{Action}, \texttt{StateAction}, \texttt{StateDiff} at different option lengths $K$. Dotted lines indicate $K=1$, solid lines indicate $K=3$, and dashed lines indicate $K=5$. $K=3$ was used across all environments for the results in the main text.}
    \label{fig:option length ablation}
\end{figure}
\begin{figure}
    \centering
    \includegraphics[width=\textwidth]{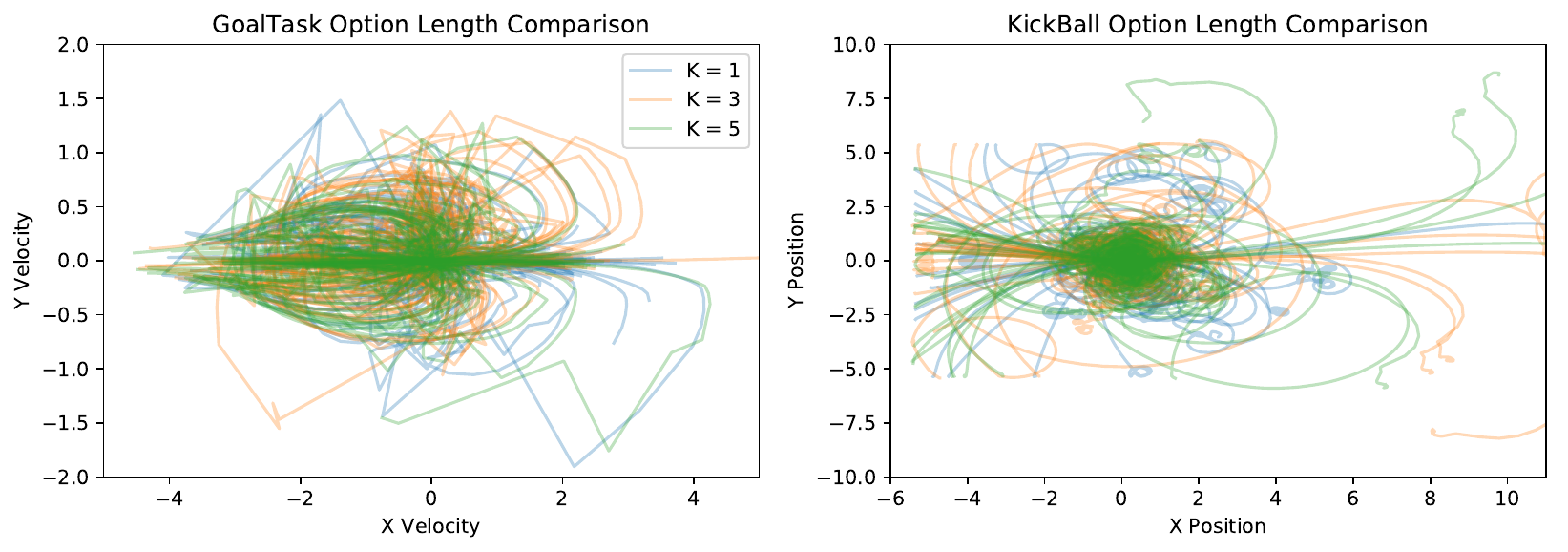}
    \caption{Trajectory distributions compared for different option lengths $K$ for the \texttt{StateAction} \method\ instantiation in both \textsc{SocialRobot} environments. These are obtained by randomly sampling an option uniformly in $[-1, 1]^D$ and keeping it fixed for the entire trajectory. 100 trajectories from each option are visualized and plotted in different colors.}
    \label{fig:option length trajectory vis}
\end{figure}
We ablate the option length $K$ in all four environments 
on the three best \method\ instantiations in Figure~\ref{fig:option length ablation}.
$K=\{1, 3, 5\}$ timesteps per option are shown, with $K=3$
and $K=5$ performing similarly across all environments,
but $K=1$ performing very poorly in comparison. $K=1$ provides no temporal abstraction, resulting in worse sample efficiency in \textsc{Pusher} and \textsc{Reacher}, and failing to learn in \textsc{GoalTask} and \textsc{KickBall}.
Although $K=5$ and $K=3$ are generally similar,
we see in \textsc{GoalTask} that $K=5$ results in better performance than $K=3$ across all three instantiations, demonstrating the potential benefit of
longer temporal abstraction lengths. 

We also plot the distribution of $(x, y)$ velocities\footnote{Velocities are relative to the agent's yaw rotation. Because \textsc{GoalTask} has egocentric inputs, the agent is not aware of the absolute $(x, y)$ coordinates in this task.} in \textsc{GoalTask} and $(x, y)$ coordinates in \textsc{KickBall} of randomly sampled options of different lengths in Figure~\ref{fig:option length trajectory vis}. Despite the fact that these two dimensions only represent a small
subspace of the entire (30-dimensional) state space, they still demonstrate
a difference in option behavior at different option lengths. We
can see that as the option length $K$ increases, the option behaviors become more consistent within a trajectory.
Meanwhile regarding coverage, $K=1$'s {\color[rgb]{0.12156862745098039, 0.4666666666666667, 0.7058823529411765}(blue)} trajectory distribution in both environments is less concentrated near the center, while 
$K=5$ {\color[rgb]{0.17254901960784313, 0.6274509803921569, 0.17254901960784313} (green)} is the most concentrated at the center. $K=3$ {\color[rgb]{1.0, 0.4980392156862745, 0.054901960784313725}(orange)} lies somewhere in between. 
We believe that this difference in behavior signifies a trade off between the coverage of the state space
and how consistent the learned options can be depending on the option length. Given the same entropy coefficient ($\beta$ in Eq~\ref{eq:worker_objective}), with longer option lengths,
it is likely that the discriminator can more easily discriminate the sub-trajectories 
created by these options, so that their coverage does not have to be as wide for the worker policy
to obtain high intrinsic rewards. Meanwhile, with shorter option lengths, 
the shorter sub-trajectories have to be more distinct for the discriminator
to be able to successfully differentiate between the options.

%This implies that in addition to improved temporal abstractions, longer
%\method\ option lengths can result in learned options that are tailored even more specifically to the
%task being solved.

\subsection{Option Visualizations}
\label{sec:extra skill visualizations}
\begin{figure}
    \centering
    \includegraphics[width=1.0\textwidth]{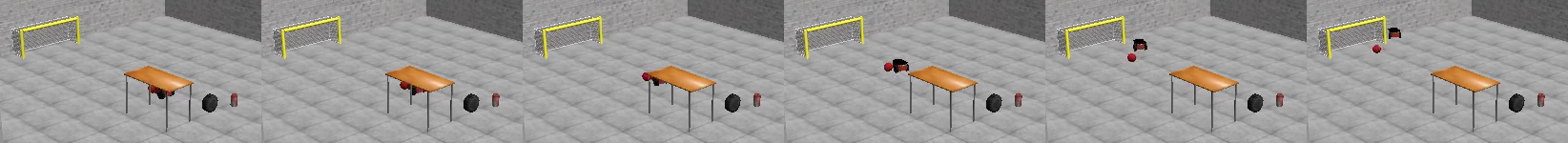}
    \includegraphics[width=1.0\textwidth]{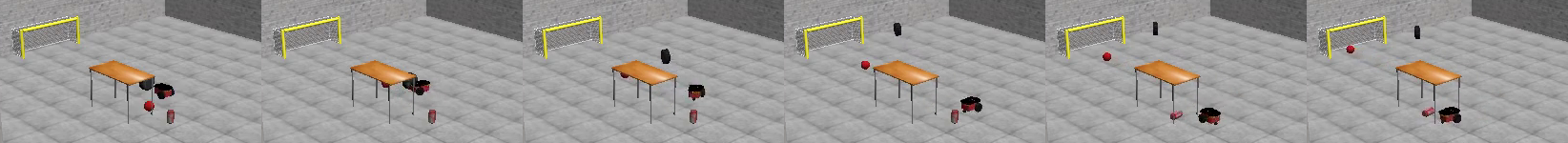}
    \includegraphics[width=1.0\textwidth]{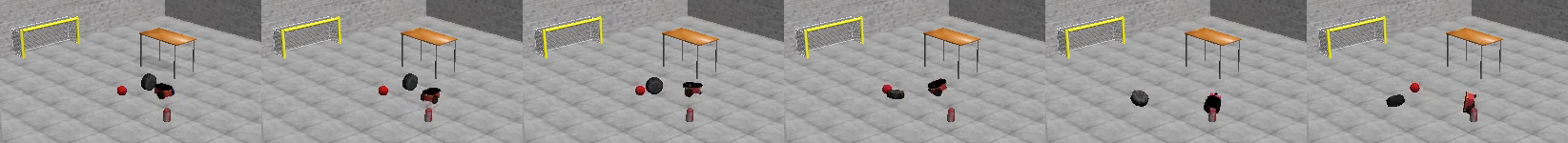}
    \includegraphics[width=1.0\textwidth]{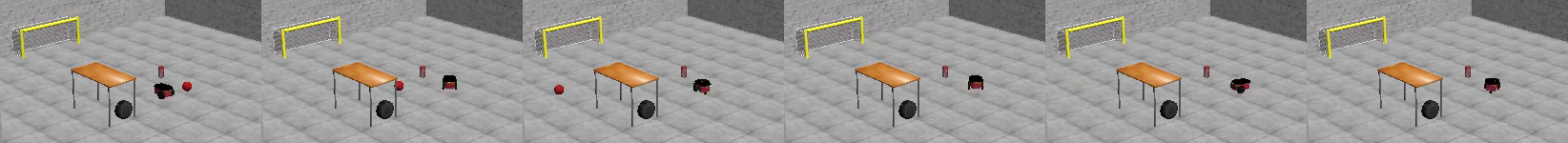}
    \includegraphics[width=1.0\textwidth]{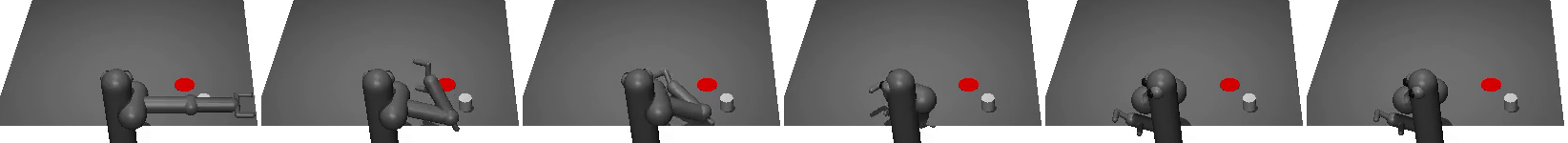}
    \includegraphics[width=1.0\textwidth]{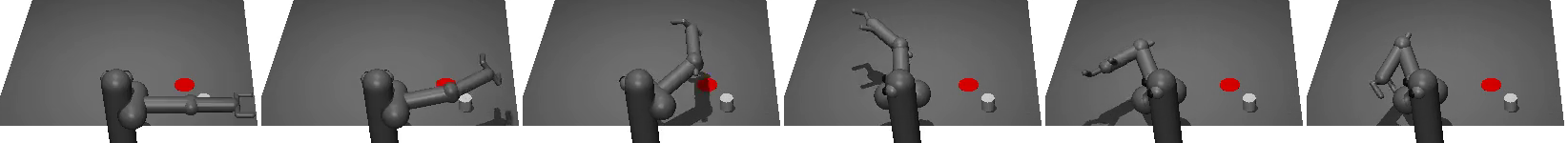}
    \includegraphics[width=1.0\textwidth]{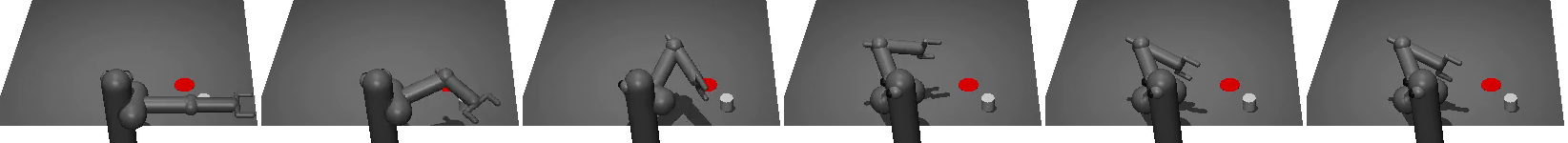}
    \includegraphics[width=1.0\textwidth]{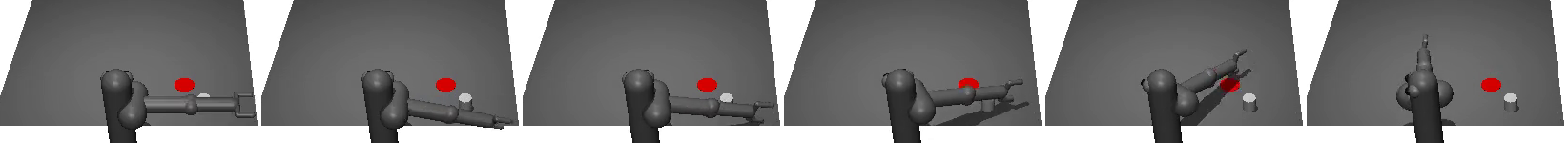}
    \caption{Eight example options from the \texttt{StateAction}
    instantiation on \textsc{KickBall} (top 4) and \textsc{Pusher}
    (bottom 4).}
    \label{fig:extra skill visualization}
\end{figure}

We visualize more option behaviors in Figure~\ref{fig:extra skill visualization}, produced in the same way as in Figure~\ref{fig:skill visualization} and as detailed in Section~\ref{sec:skill visualizations}.
The top 4 picture reels are from \textsc{KickBall}.
We see that \textsc{KickBall} options lead to varied
directional driving behaviors that can be utilized
for efficient navigation. For example, the second, third, and fourth highlight
options that produce right turning behavior, however 
at different speeds and angles.
The option in the third reel is a quick turn that
results in the robot tumbling over into an 
unrecoverable state, but the options in the second 
and fourth reels
turn more slowly and do not result in the
robot flipping. The first option simply proceeds forward
from the robot starting position, kicking the ball into
the goal.

The bottom 4 reels are from \textsc{Pusher}. Each option results in different sweeping behaviors with 
varied joint positioning and arm height. These sweeping and arm folding behaviors, when utilized in short sub-trajectories, are useful for controlling where 
and how to move the arm to push the puck into the goal.

\section{Hyperparameters}
\label{sec:hyperparameters} 
To ensure a fair
comparison across all methods, we perform a hyperparameter search over the
following values for each algorithm and suite of environments.

\subsection{\textsc{Pusher} and \textsc{Reacher}} 
\label{sec:7-DOF hyperparams}
Shared hyperparameters across all methods are listed below (where applicable, and
except when overridden by hyperparameters listed for each individual method).
For all methods, we take the hyperparameters
that perform best across 3 random seeds in terms of the area under the
evaluation success curve (AUC) in the \textsc{Pusher} environment.

\begin{itemize}
    \item Number of parallel actors/environments per rollout: 20 
    \item Steps per episode: 100
    %\item Environment steps per rollout: 2000 ($20 \times 100$) \haonan{This depends on rollout length, not episode length}
    \item Batch size: 2048 
    \item Learning rate: $10^{-4}$ for all network modules
    \item Policy/Q network hidden layers: (256, 256, 256) with ReLU non-linearities
    \item Polyak averaging coefficient for target Q: 0.999
    \item Target Q update interval (training iterations): 1
    \item Training batches per iteration: 100
    \item Episodes per evaluation: 50
    \item Initial environment steps for data collection before training: 10000
\end{itemize}
Rollouts and training iterations are performed alternatively, one after the
other. The \emph{rollout length} searched below refers to how many time steps in each environment are taken per rollout/training iteration, 
effectively controlling the ratio of gradient steps to environment steps. 
A smaller rollout length corresponds to a higher ratio.
This ratio is also searched over for \textsc{HiPPO} and \textsc{HIRO}.
Other hyperparameters searched separately for each algorithm are listed
below, and selected ones are \textbf{bolded}.

\subsubsection{SAC}
\begin{itemize}
    \item Target entropy min prob $\Delta$\footnote{The target entropy used for
    automatically adjusting $\alpha$ is calculated as: $\sum_{i} [\ln (M_i - m_i)
    + \ln \Delta]$ where $M_i$/$m_i$ are the maximium/minimum
    value of action dim $i$. Intuitively, the target distribution concentrates on a 
    segment of length $(M_i-m_i)\Delta$ with a constant probability.}: \{0.1, \textbf{0.2}, 0.3\}
    \item Replay buffer length per parallel actor: \{50000, \textbf{200000}\}
    \item Rollout Length: \{12, 25, \textbf{50}, 100\}
\end{itemize}
\subsubsection{SAC w/ Action Repetition}
\begin{itemize}
    \item Action repetition length\footnote{Chosen to match the option interval $K$ of \method.}: 3
    \item Rollout Length: \{\textbf{4}, 8, 16, 33\}
\end{itemize}
Other hyperparameters are kept the same as the optimal SAC ones.

\subsubsection{\method}
The hyperparameters of \method\ were mostly heuristically chosen due to the
hyperparameter search space being too large.
\begin{itemize}
    \item Latent option $\bfu$ vector dimension ($D$): \{\textbf{8}, 12\}
    \item Policy/Q network hidden layers for \pilow\ : (128, 128, 128)
    \item Steps per option ($K$): 3
    \item \pilow\ has a fixed entropy coefficient $\alpha$ of 0.01. Target entropy min prob $\Delta$ for \pihi\ is 0.2.
    \item Discriminator network hidden layers: (64, 64)
    \item Replay buffer length per parallel actor: \{50000, \textbf{200000}\}
    \item Rollout Length: \{\textbf{25}, 50, 100\}
\end{itemize}

\subsubsection{HIRO}
\begin{itemize}
    \item Steps per option: \{\textbf{3}, 5, 8\}
    \item Replay buffer size (total): \{500000, \textbf{2000000}\}
    \item Meta action space (actions are relative, \eg\ meta-action is
    \texttt{current\_obs + action}): \texttt{(-np.ones(obs\_space - 3D\_goal\_pos)*2,
    np.ones(obs\_space - 3D\_goal\_pos)*2})
    \item Policy stddev noise: \{\textbf{0.1}, 0.3, 0.5\}
    \item Number of gradient updates per training iteration: \{100, \textbf{200}, 400\}
\end{itemize}

\subsubsection{HiPPO}
For most hyperparameters, the search ranges chosen were derived after discussion with the first author of HiPPO.
\begin{itemize}
    \item Learning rate: \textbf{$3\times 10^{-4}$}
    \item Policy network hidden layers: (256, 256)
    \item Skill selection network hidden layers: \{(32, 32), \textbf{(128, 64)}\}
    \item Latent skill vector size: \{5, \textbf{10}, 15\}
    \item PPO clipping parameter: \{0.05, \textbf{0.1}\}
    \item Time commitment range: \{(2, 5), \textbf{(3, 7)}\}
    \item Policy training steps per epoch: \{\textbf{25}, 50, 100\}
\end{itemize}

\subsection{\textsc{SocialRobot}} 
For all methods, we select the hyperparameters with the best
area under the evaluation success curve (AUC) in the \textsc{KickBall}
environment, and apply them to both \textsc{KickBall} and \textsc{GoalTask}.
The shared hyperparameters are as follows (if
applicable to the algorithm, and except when overridden by the respective
algorithm's list of hyperparameters):
\begin{itemize}
    \item Number of parallel actors/environments per rollout: 10
    \item Steps per episode: 100 (\textsc{GoalTask}), 200 (\textsc{KickBall})
    %\item Environment steps per rollout: 1000 ($100 \times 10$)
    \item Batch size: 1024
    \item Learning rate: $5\times 10^{-4}$ for all network modules
    \item Policy/Q network hidden layers: (256, 256, 256) with ReLU
    non-linearities
    \item Polyak averaging coefficient for target Q: 0.95
    \item Target Q update interval (training iterations): 1
    \item Training batches per iteration: 100
    \item Episodes per evaluation: 100
    \item Evaluation interval (training iterations): 100
    \item Initial environment steps for data collection before training:
    100000
\end{itemize}
The training terminology here generally follows section~\ref{sec:7-DOF hyperparams}.

\subsubsection{SAC}
\begin{itemize}
    \item Target entropy min prob $\Delta$: \{0.1, \textbf{0.2}, 0.3\}
    \item Replay buffer length per parallel actor: \{20000, \textbf{100000}\}
    \item Rollout length: \{12, 25, \textbf{50}, 100\}
\end{itemize}

\subsubsection{SAC w/ Action Repetition}
\begin{itemize}
    \item Action repetition length\footnote{Chosen to match the option interval $K$ of \method.}: 3
    \item Rollout Length: \{4, 8, 16, \textbf{33}\}
\end{itemize}
Other hyperparameters are kept the same as the optimal SAC ones.

\subsubsection{\method}
Due to the large hyperparameter search space, we only search over the option
vector size and rollout length, and select everything else heuristically.
\begin{itemize}
    \item Latent option $\bfu$ vector dimension ($D$): \{\textbf{4}, 6\}
    \item Policy/Q network hidden layers for \pilow\: (128, 128, 128)
    \item Steps per option ($K$): 3
    \item \pilow\ has a fixed 
    entropy coefficient $\alpha$ of 0.01. Target entropy min prob $\Delta$ for \pihi\ is 0.2.
    \item Discriminator network hidden layers: (32, 32)
    \item Replay buffer length per parallel actor: 20000
    \item Rollout Length: \{50, \textbf{100}\}
\end{itemize}
\subsubsection{HIRO}
\begin{itemize}
    \item Learning rate: $3\times 10^{-4}$
    \item Steps per option: \{\textbf{3}, 5, 8\}
    \item Replay buffer size (total): \{\textbf{500000}, 2000000\}
    \item Meta action space (actions are relative, \eg\ meta-action is
    \texttt{current\_obs + action}):
    \begin{itemize}
        \item \textsc{GoalTask}: \texttt{(-np.ones(obs\_space) * 2,
        np.ones(obs\_space) * 2)}
        \item \textsc{KickBall}: \texttt{(-np.ones(obs\_space -
        goal\_space) * 2, np.ones(obs\_space - goal\_space) * 2)} (because
        the goal position is given but will not change in the observation
        space)
    \end{itemize}
    \item Policy stddev noise \{\textbf{0.1}, 0.3, 0.5\}
    \item Number of gradient updates per training iteration: \{100, 200, \textbf{400}\}
\end{itemize}
\subsubsection{HiPPO}
\begin{itemize}
    \item Learning rate: $3\times 10^{-4}$
    \item Policy network hidden layers: \{\textbf{(64, 64)}, (256, 256)\}
    \item Skill selection network hidden layers: \{\textbf{(32, 32)}, (128,
    64)\}
    \item Latent skill vector size: \{4, \textbf{8}\}
    \item PPO clipping parameter: \{\textbf{0.05}, 0.1\}
    \item Time commitment range: \{\textbf{(2, 5)}, (3, 7)\}
    \item Policy training steps per epoch: \{25, \textbf{50}, 100\}
\end{itemize}
\end{document}